%% file: main_arxiv.tex
\documentclass{article}

\usepackage{nac_preprint}

\usepackage[utf8]{inputenc} 
\usepackage[T1]{fontenc}    
\usepackage{hyperref}       
\usepackage{url}            
\usepackage{booktabs}       
\usepackage{amsfonts}       
\usepackage{nicefrac}       
\usepackage{microtype}      
\usepackage{xcolor}         

\usepackage{booktabs}
\usepackage{subcaption}
\usepackage{amsmath}
\usepackage{mathtools}
\usepackage[toc,page]{appendix}
\usepackage{enumitem}
\usepackage{graphics}
\usepackage{graphicx}
\usepackage[export]{adjustbox}
\usepackage{algorithm}
\usepackage{algorithmicx}
\usepackage[noend]{algpseudocode}
\algnewcommand{\LineComment}[1]{\State \(//\) #1}

\usepackage{wrapfig}

\title{Backprop-Free Reinforcement Learning with \\Active Neural Generative Coding}
\author{%
  Alexander Ororbia$^\dagger$ \\
  \texttt{ago@cs.rit.edu} \\
  \And
  Ankur Mali$^\ddagger$ \\
  \texttt{aam35@ist.psu.edu} 
  \AND
  \vspace{-0.75cm}\\
  $^\dagger$ Rochester Institute of Technology, Rochester, NY 14623 \\
  $^\ddagger$ Pennsylvania State University, University Park, PA 16802
}

\begin{document}

\maketitle

\begin{abstract}
 In humans, perceptual awareness facilitates the fast recognition and extraction of information from sensory input. This awareness largely depends on how the human agent interacts with the environment. In this work, we propose \emph{active neural generative coding}, a computational framework for learning action-driven generative models without backpropagation of errors (backprop) in dynamic environments. Specifically, we develop an intelligent agent that operates even with sparse rewards, drawing inspiration from the cognitive theory of planning as inference. We demonstrate on several simple control problems that our framework performs competitively with deep Q-learning. The robust performance of our agent offers promising evidence that a backprop-free approach for neural inference and learning can drive goal-directed behavior.
\end{abstract}


\section{Introduction}
\label{sec:intro}

Manipulating one's environment in the effort to understand it is an essential ingredient of learning in humans \cite{spielberger1994curiosity,berlyne1966curiosity}. In cognitive neuroscience, behavioral and neurobiological evidence indicates a distinction between goal-directed and habitual action selection in reward-based decision-making. With respect to habitual action selection, or actions taken based on situation-response associations, ample evidence exists to support the temporal-difference (TD) account from reinforcement learning. In this account, the neurotransmitter dopamine creates an error signal based on reward prediction to drive (state) updates in the corpus striatum, a particular neuronal region in the basal ganglia that affects an agent's choice of action. In contrast, goal-directed action requires prospective planning where actions are taken based on predictions of their future potential outcomes \cite{niv2009reinforcement,solway2012goal}. Planning-as-inference (PAI) \cite{botvinick2012planning} attempts to account for goal-directed behavior by casting it as a problem of probabilistic inference where an agent manipulates an internal model that estimates the probability of potential action-outcome-reward sequences.

One important, emerging theoretical framework for PAI is that of active inference \cite{friston2011action,tschantz2020learning}, which posits that biological agents learn a probabilistic generative model by interacting with their world, adjusting the internal states of this model to account for the evidence that they acquire from their environment. This scheme unifies perception, action, and learning in adaptive systems by framing them as processes that result from approximate Bayesian inference, elegantly tackling the exploration-exploitation trade-off inherent to organism survival.
The emergence of this framework is timely -- in reinforcement learning (RL) research, despite the recent successes afforded by artificial neural networks (ANNs) \cite{mnih2013playing,silver2018general}, most models require exorbitant quantities of data to train well, struggling to learn tasks as efficiently as humans and animals \cite{arulkumaran2017brief}. As a result, a key challenge is how to design RL methods that successfully resolve environmental uncertainty and complexity given limited resources and data. 
Model-based RL, explored in statistical learning research through world \cite{ha2018recurrent} or dynamics models \cite{sutton1990integrated}, offers a promising means of tackling this challenge and active inference provides a promising path towards instantiating it in a powerful yet neurocognitively meaningful way \cite{tschantz2020reinforcement}.


Although PAI and active inference offer an excellent story for biological system behavior and a promising model-based RL setup, most computational implementations are formulated with explainability in mind (favoring meaningfully labeled albeit low-dimensional, discrete state/action spaces) yet in the form of complex probabilistic graphical models that do not scale easily \cite{friston2015active,friston2017active,friston2017active_curosity,friston2018deep}. In response, effort has been made to scale active inference by using deep ANNs \cite{ueltzhoffer2018deep,tschantz2020scaling} trained by the popular backpropagation of errors (backprop) \cite{rumelhart_learning_1986}. While ANNs represent a powerful step in the right direction, one common criticism of using them within the normative framework of RL is that they have little biological relevance despite their conceptual value \cite{lake2017building,zador2019critique}. Importantly, from a practical point-of-view, they also suffer from practical issues related to their backprop-centric design \cite{ororbia2018biologically}. 
This raises the question: can a biologically-motivated alternative to backprop-based ANNs also facilitate reinforcement learning through active inference in a scalable way?
In this paper, motivated by the fact that animals and humans solve the RL problem, we develop one alternative that positively answers this question.
As a result, the neural agent we propose represents a promising step forward towards better modeling the approximations that biological neural circuitry implements when facing real-world resource constraints and limitations, creating the potential for developing new theoretical insights.
Such insights will allow us to design agents better capable of dealing with continuous, noisy sensory patterns \cite{niv2009reinforcement}.
 
\begin{figure}[!t]
    \centering
    \includegraphics[width=0.775\textwidth]{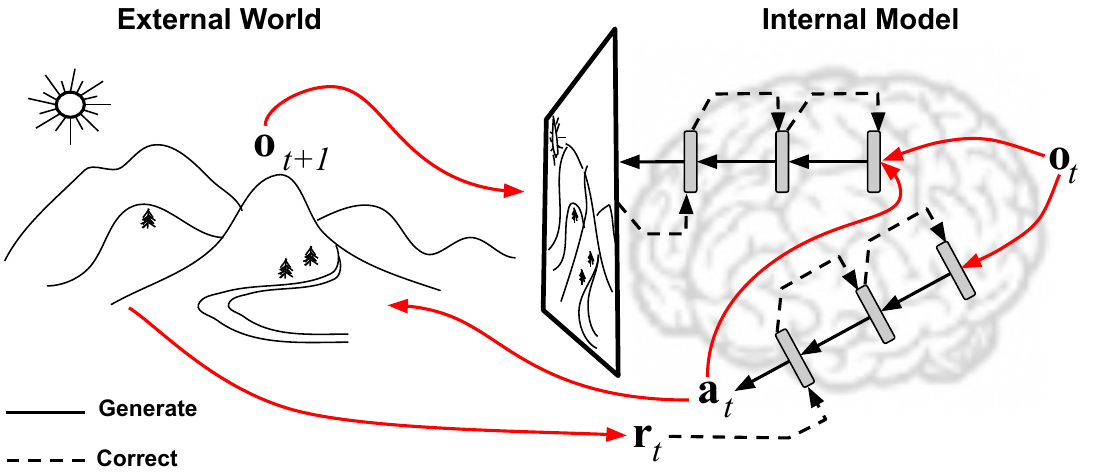}
    \caption{
    An ANGC agent predicts the world, acts to manipulate it, and then corrects itself given observations/rewards. }
    \label{fig:angc_intuition}
\end{figure}

While many backprop-alternative (backprop-free) algorithms have recently been proposed \cite{movellan1991contrastive,o1996biologically,lee2015difference,lillicrap2016random,scellier2017equilibrium,guerguiev2017towards,whittington2017approximation,ororbia2018biologically}, few have been investigated outside the context of supervised learning, with some notable exceptions in sequence \cite{wiseman2017training,ororbia2020continual,manchev2020target} and generative modeling \cite{ororbia2020neural}. In the realm of RL, aside from neuro-evolutionary approaches \cite{deepneuro,heidricheisner2009neuro} or methods that build on top of them \cite{najarro2020meta}, the dearth of work is more prescient and our intent is to close this gap by providing a backprop-free approach to inference and learning, which we call \emph{active neural generative coding} (ANGC), to drive goal-oriented agents. In our system, we demonstrate how a scalable, biologically-plausible inference and learning process, grounded in the theory of predictive processing \cite{friston2005theory,clark2015surfing}, can lead to adaptive, self-motivated behavior in the effort to balance the exploration-exploitation trade-off in RL.
One key element to consider is that ANGC offers robustness in settings with sparse rewards which other backprop-free methods such as neuroevolution \cite{deepneuro,heidricheisner2009neuro} struggle with.\footnote{It is difficult to determine a strong encoding scheme as well as an effective breeding strategy for the underlying genetic algorithm. Such design choices play a large role in the success of the approach.}
To evaluate ANGC's efficacy, we implement an agent structure tasked with solving control problems often experimented with in RL and compare performance against several backprop-based approaches.


\section{Active Neural Generative Coding}
\label{sec:angc}

To specify our proposed ANGC agent, the high-level intuition of which is illustrated in Figure \ref{fig:angc_intuition}, we start by defining of the fundamental building block used to construct it, i.e., the neural generative coding circuit. Specifically, we examine its neural dynamics (for figuring out hidden layer values given inputs and outputs) and its synaptic weight updates.

\subsection{The Neural Generative Coding Circuit}
\label{sec:ngc}

Neural generative coding (NGC) is a recently developed framework \cite{ororbia2020neural} that generalizes classical ideas in predictive processing \cite{rao1999predictive,clark2015surfing} to the construction of scalable neural models that model and predict both static and temporal patterns \cite{ororbia2020continual,ororbia2020neural}. An NGC model is composed of $L+1$ layers of stateful neurons  $\mathfrak{N}^0,\mathfrak{N}^1,\cdots,\mathfrak{N}^L$ that are engaged in a process of never-ending guess-then-correct, where $\mathfrak{N}^\ell$ contains $J_\ell$ neurons (each neuron has a latent state value represented by a scalar). 
The combined latent state of the neurons in $\mathfrak{N}^\ell$ is represented by the vector $\mathbf{z}^\ell \in \mathcal{R}^{J_\ell \times 1}$ (initially $\mathbf{z}^\ell = \mathbf{0}$ in the presence of a new data pattern).
Generally, an NGC model's bottom-most layer $\mathfrak{N}^0$ is clamped to a sensory pattern extracted from the environment. However, in this work, we design a model that clamps both its top-most layer $\mathfrak{N}^L$ and bottom-most layer $\mathfrak{N}^0$ to particular sensory variables, i.e., $\mathbf{z}^L = \mathbf{x}^i$ and $\mathbf{z}^0 = \mathbf{x}^o$, allowing the agent to process streams of data $(\mathbf{x}^i, \mathbf{x}^o)$ where $\mathbf{x}^i \in \mathcal{R}^{J_L \times 1}$ and $\mathbf{x}^o \in \mathcal{R}^{J_0 \times 1}$. 

\begin{figure*}
\centering
\begin{subfigure}[b]{0.4\linewidth}
\begin{center}
\includegraphics[width=0.435\linewidth]{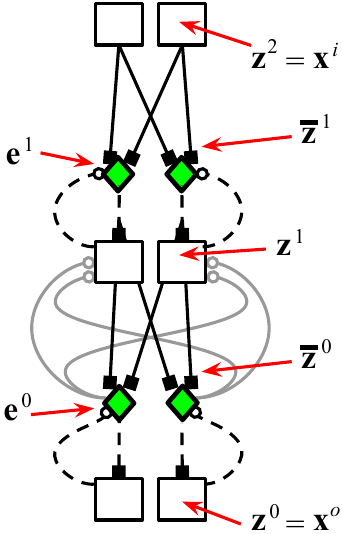}
\end{center}
\label{fig:ngc_circuit}
\end{subfigure}
\begin{subfigure}[b]{0.565\linewidth}
\begin{center}
\includegraphics[width=0.83\linewidth]{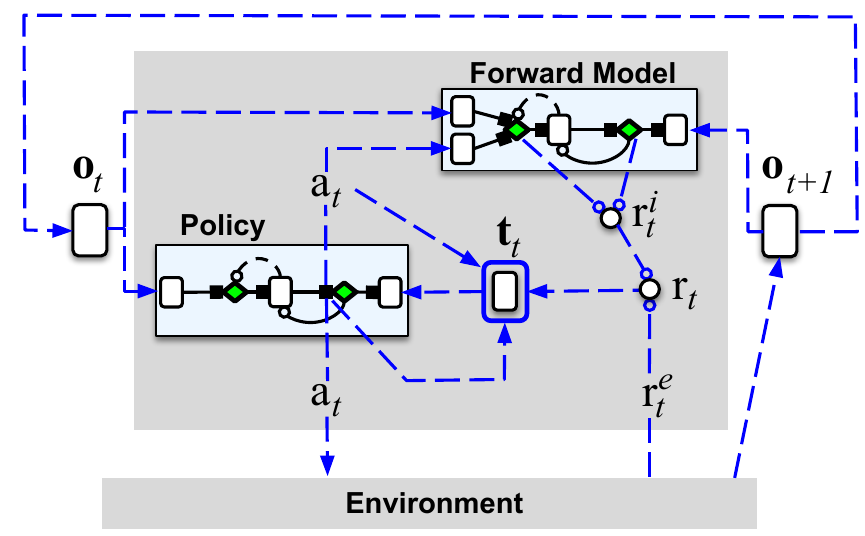}
\end{center}
\label{fig:angc_agent}
\end{subfigure}
\caption{The NGC circuit (left) and the high-level ANGC architecture, with both a controller and generator, (right). Green diamonds represent error neurons, empty rectangles represent state neurons, solid arrows represent individual synapses, dashed arrows represent direct copying of information, open circles indicate excitatory signals, and filled squares indicate inhibitory signals. Note that $\mathbf{t}_t$, which is enclosed in a rounded blue box, entails a more intricate calculation (see Equation \ref{eqn:target_vector}). }
\label{fig:angc}
\end{figure*}

Specifically, in an NGC model, layer $\mathfrak{N}^{\ell+1}$   
attempts to guess the current post-activity values of layer $\mathfrak{N}^\ell$, i.e., $\phi^\ell(\mathbf{z}^{\ell})$, by generating a prediction vector $\mathbf{\bar{z}}^\ell$ using a matrix of forward synaptic weights $\mathbf{W}^{\ell+1} \in \mathcal{R}^{J_{\ell} \times J_{\ell+1}}$. The prediction vector is then compared against the target activity by a corresponding set of error neurons $\mathbf{e}^\ell$ which simply perform a direct mismatch calculation as follows:
$\mathbf{e}^{\ell} = \frac{1}{\beta_e}(\phi^\ell( \mathbf{z}^\ell ) - \mathbf{\bar{z}}^\ell)$.\footnote{One may replace $\frac{1}{2\beta_e}$ with $\Sigma^{-1}$, i.e., a learnable matrix that applies precision-weighting to the error units, as in \cite{ororbia2020neural}. We defer using this scheme for future work.} 
This error signal is finally transmitted back to the layer that made the prediction $\mathbf{\bar{z}}^\ell$ through a complementary matrix of error synapses $\mathbf{E}^{\ell+1} \in \mathcal{R}^{J_{\ell+1} \times J_{\ell}}$. Given the description above, the set of equations that characterize the NGC neural circuit and its key computations are:
\begin{align}
    \mathbf{\bar{z}}^\ell &= g_\ell \Big( \mathbf{W}^{\ell+1} \cdot \phi^{\ell+1}(\mathbf{z}^{\ell+1}) \Big), \;  \mathbf{e}^\ell = \frac{1}{2\beta_e} (\phi^\ell( \mathbf{z}^\ell ) - \mathbf{\bar{z}}^\ell) \label{eqn:ngc_predict} \\
     \mathbf{z}^{\ell+1} &\leftarrow \mathbf{z}^{\ell+1} + \beta \Big( \overbrace{-\gamma_v \mathbf{z}^{\ell+1}}^\text{leak} + \overbrace{\mathbf{d}^{\ell+1}}^{\text{pressure}} + \overbrace{\vartheta(\mathbf{z}^{\ell+1})}^\text{lateral term} ) \Big)  \label{eqn:ngc_correct} \\
     &\mbox{where} \; \mathbf{d}^{\ell+1} = -\mathbf{e}^{\ell+1} + ( \mathbf{E}^{\ell+1} \cdot \mathbf{e}^\ell) \nonumber 
\end{align}
where $\cdot$ indicates matrix/vector multiplication and $\phi^{\ell+1}$ and $g_\ell$ are element-wise activation functions, e.g., the hyperbolic tangent $tanh(v) = (\exp(2v) - 1)/(\exp(2v) + 1)$ or the linear rectifier $\phi^\ell(v) = max(0, v)$. In this paper, we set $g_\ell$ as the identity, i.e., $g_\ell(v) = v$, for all layers.
In Equation \ref{eqn:ngc_correct}, the coefficient that weights the correction applied to state layer $\ell+1$ is determined by the formula $\beta = \frac{1}{\tau}$ where $\tau$ is the integration time constant in the order of milliseconds.
The leak variable $-\gamma_v \mathbf{z}^\ell$ decays the state value over time ($\gamma_v$ is a positive coefficient to control its strength). $\vartheta(\mathbf{z}^\ell)$ is lateral excitation/inhibition term, which is a function that creates competition patterns among the neurons inside of layer $\mathfrak{N}^\ell$ \cite{ororbia2020neural} -- in this paper we set $\vartheta(\mathbf{z}^\ell) = 0$ since its effect is not needed for this study. Upon encountering data $(\mathbf{x}^i, \mathbf{x}^o)$, the model's top and bottom layers are clamped, i.e., $\mathbf{z}^L = \mathbf{x}^i$ and $\mathbf{z}^0 = \mathbf{x}^o$, and Equations \ref{eqn:ngc_predict}-\ref{eqn:ngc_correct} are run $K$ times in order to search for activity values $\{ \mathbf{z}^1,\cdots,\mathbf{z}^{L-1} \}$ (see $\Call{Infer}{}$ in the pseudocode provided in the appendix).

After the internal activities have been found, the synaptic weights may be adjusted using a modulated local error Hebbian rule adapted from local representation alignment (LRA) \cite{ororbia2018biologically,ororbia2020large}:
\begin{align}
    \Delta \mathbf{W}^\ell &= \mathbf{e}^{\ell} \cdot (\phi^\ell(\mathbf{z}^{\ell+1}))^T \otimes \mathbf{M}^\ell_W \label{eqn:forward_update} \\
    \Delta \mathbf{E}^\ell &= \gamma_e (\Delta \mathbf{W}^\ell)^T \otimes \mathbf{M}^\ell_E \label{eqn:backward_update}
\end{align}
where $\gamma_e$ controls the time-scale at which the error synapses are adjusted (usually values in the range of $[0.9,1.0]$ are used). The dynamic modulation factors $\mathbf{M}^\ell_W$ and $\mathbf{M}^\ell_E$ help to stabilize  learning in the face of non-stationary streams and are based on insights related to nonlinear synaptic dynamics (see Appendix for a full treatment of these factors).
Once the weight adjustments have been computed, an update rule such as stochastic gradient ascent, Adam \cite{kingma2014adam}, or RMSprop \cite{Tieleman2012} can be used (see $\Call{UpdateWeights}{}$ in Algorithm CC, Appendix).

The online objective that an NGC model attempts to minimize is known as total discrepancy (TotD) \cite{ororbia2017learning}, from which the error neuron, state update expressions, and local synaptic adjustments may be derived \cite{ororbia2020large,ororbia2020neural}. The (TotD) objective, which could also be interpreted as a form of free energy \cite{friston2009free} specialized for stateful neural models that utilize arbitrary forward and error synaptic pathways \cite{ororbia2020large}, can be expressed in many forms including the linear combination of local density functions \cite{ororbia2020neural} or the summation of distance functions \cite{ororbia2020continual}. 
For this study, the form of TotD used is:
\begin{align*}
    \mathcal{L}(\Theta) = \sum^{L-1}_{\ell=0} \mathcal{L}(\mathbf{z}^\ell, \mathbf{\bar{z}}^\ell) = \sum^{L-1}_{\ell=0} \frac{1}{\beta_e}||\phi^\ell(\mathbf{z}^\ell) - \mathbf{\bar{z}}^\ell||^2_2 \mbox{.}
\end{align*}

Algorithm CC (see Appendix), puts all of the equations and relevant details presented so far together to describe inference and learning under a full NGC model that processes $(\mathbf{x}^i, \mathbf{x}^o)$ from a data stream. We note here that the algorithm breaks down model processing into three routines -- $\Call{Infer}{\circ}$, $\Call{UpdateWeights}{\circ}$, and $\Call{Project}{\circ}$. $\Call{Infer}{\circ}$ is simply the $K$-step described earlier to find reasonable values of the latent state activities given clamped data and $\Call{UpdateWeights}{\circ}$ is the complementary procedure used to adjust the synaptic weight parameters once state activities have been found after using $\Call{Infer}{\circ}$. $\Call{Project}{\circ}$ is a special function that specifically clamps data $\mathbf{x}^i$ to the top-most layer and projects this information directly through the underlying directed graph defined by the NGC architecture -- this routine is essentially a variant of the ancestral sampling procedure defined in \cite{ororbia2020neural} but accepts a clamped input pattern instead of samples drawn from a prior distribution.
Figure \ref{fig:angc} (left) graphically depicts a three layer NGC model with $2$ neurons per layer.

\subsection{Generalizing to Active Neural Coding}
\label{sec:active_coding}

Given the definition of the NGC block, we turn our attention to its generalization that incorporates actions, i.e., active NGC (ANGC). ANGC is built on the premise that an agent adapts to its environment by balancing a trade-off between (at least) two key quantities -- surprisal and preference. This means that our agent is constantly tracking a measurement of how surprising the observations it encounters are at a given time step (which drives exploration) as well as a measurement of its progress towards a goal. In effect, maximizing the sum of these two terms means that the agent will seek observations that are most ``suprising'' (which yield the most information when attempting to reduce uncertainty) while attempting to reduce its distance to a goal state (which maximizes the discounted long-term future reward).
Formally, this means that an ANGC agent will maximize:
\begin{align}
    r_t = \alpha_e r^e_t + \alpha_i r^i_t = r^{in}_t + r^{ep}_t \label{eqn:reward}
\end{align}
which is a reward signal that can be decomposed into an instrumental (or goal-oriented) signal $r^{in}_t$ and an epistemic (or exploration/information maximizing) signal $r^{ep}_t$. Each component signal is controlled by an importance factor, $\alpha_e$ for the epistemic term and $\alpha_i$ for the instrumental term, and a raw internal signal produced either by the generative model ($r^e_t$ to drive $r^{ep}_t$) or an external goal-directing signal ($r^i_t$ to drive $r^{in}_t$).
%
Although we chose to interpret and represent the active inference view of the exploration-exploitation trade-off as (dopamine) scalars, our instrumental signal is not limited to this scheme and could incorporate an encoding of more complex functions such as (prior) distribution functions over (goal) states (see Appendix).

%

As indicated by the architecture diagram in Figure \ref{fig:angc} (right), the implementation of our ANGC agent in this paper is a coupling of two NGC circuits -- the generator (or dynamic generative model), which is responsible for producing the epistemic term $r^{ep}_t$, and the controller, which is responsible for choosing the actions such that the full reward $r_t$, which includes the instrumental term $r^{in}_t$, is maximized.

\paragraph{The NGC Generator (Forward Model)} Once the generator's top-most latent state is clamped to the current $D$-dimensional observation $\mathbf{o}_t \in \mathcal{R}^{D \times 1}$ and the 1-of-$A$ encoding of the controller's currently chosen action $a_t$ (out of $A$ possible actions), i.e., $\mathbf{a}_t \in \{0,1\}^{A \times 1}$, the generator attempts to predict the value of the next observation of the environment $\mathbf{o}_{t+1}$. Using the routine $\Call{Infer}{}$ (Algorithm CC, Appendix), the generator, with parameters $\Theta_g$, searches for a good set of latent state activities to explain output $\mathbf{x}^o = \mathbf{o}_{t+1}$ given input $\mathbf{x}^i = [\mathbf{a}_t, \mathbf{o}_{t+1}]$ where $[\cdot,\cdot]$ indicates the vector concatenation of $\mathbf{o}_t$ and $\mathbf{a}_t$. Once latent activities have been found, the generator then updates its synapses via routine $\Call{UpdateWeights}{}$ (Algorithm CC, Appendix).

The generator plays a key role in that it drives the exploration conducted by an ANGC agent. Specifically, as the generator progressively learns to how to synthesize future observations, the current activities of its error neurons embedded at each layer, i.e., $\mathcal{E} = \{\mathbf{e}^0, \mathbf{e}^1,\cdots,\mathbf{e}^L\}$, are used to produce an epistemic modulation term. Formally, this means that the exploration signal is calculated as $r^e_t = \sum_\ell ||\mathbf{e}^\ell||^2_2$ which is the result of summing across layers and across each error neuron vector's respective dimensions.\footnote{Observe that $r^e_t$ is the generator's TotD, i.e., $r^e_t = \mathcal{L}(\Theta_g)$.}
The epistemic term $r^{ep}_t = \alpha_e r^e_t$ is then combined with an instrumental term $r^{in}_t = \alpha_i r^i_t$, i.e., the scalar signal produced externally (by the environment or another neural system), to guide the agent towards a goal state(s), via Equation \ref{eqn:reward}. The final $r_t$ is then subsequently used to adapt the controller  (described next).

\paragraph{The NGC Controller (Policy Model)} With its top-most latent state clamped to the $t$th observation, i.e., $\mathbf{x}^i = \mathbf{o}_t$, the controller, with parameters $\Theta_c$, will generate a prediction of the full reward signal $r_t$. Specifically, at time $t$, given a target scalar (produced by the environment and the generator), the controller will also infer a suitable set of latent activities using the $\Call{Infer}{}$ routine defined in Algorithm CC (Appendix).

Since the NGC controller's output layer will estimate a potential reward signal for each possible discrete action that the agent could take (which is typical in many modern Q-learning setups), we must first compose the target activity $\mathbf{t}_t$ for the output nodes once the scalar value $r_t$ is obtained. This is done by first encoding the action as a 1-of-$A$ vector $\mathbf{a}_t$ (this is what done by the $\Call{toOneHot}{}$ function call in Algorithm \ref{algo:angc_process}), computing the boot-strap estimate of the future discounted reward $\mathbf{d}_{t+1} = \Call{Project}{\mathbf{o}_{t+1}, \Theta_c}$, and finally checking if the next observation is a terminal. Specifically, the target vector is computed according to the following:
\begin{align}
    \mathbf{t}_t = t_t \mathbf{a}_t + (1 - \mathbf{a}_t) \otimes \Call{Project}{\mathbf{o}_t, \Theta_c} \label{eqn:target_vector}
\end{align}
where the target scalar $t_t$ is created according to the following logical expression:
\begin{align}
    (\mathbf{o}_t \mbox{ is terminal} \rightarrow t_t = r_t) \land (\mathbf{o}_t \mbox{ is not terminal} \rightarrow t^\prime_t) \label{eqn:target_value}\\ 
    \mbox{where } \; t^\prime_t = r_t + \gamma \max_a \Call{Project}{\mathbf{o}_{t+1}, \Theta_c}) \mbox{.} \nonumber
\end{align}
Once $\mathbf{t}_t$ has been prepared, the controller is run to find its latent activities for $\mathbf{o}_t$ and $\mathbf{t}_t$ using $\Call{Infer}{}$ and calculates its local weight updates via the $\Call{UpdateWeights}{}$ routine (Algorithm CC, Appendix).
Furthermore, observe that the second sub-expression in Equation \ref{eqn:target_value} involves re-using the controller to estimate the (reward) value of the future observation, i.e., $\gamma \max_a \Call{Project}{\mathbf{o}_{t+1},\Theta_c}$ term. This term can be replaced with a proxy term $\gamma \max_a \Call{Project}{\mathbf{o}_{t+1},\widehat{\Theta}_c}$ to implement the target network stability mechanism proposed in \cite{mnih2015human}, where $\widehat{\Theta}_c$ are the parameters of a ``target controller'', initialized to be the values of $\Theta_c$ at the start of the simulation and updated every $C$ transitions by Polyak averaging $\widehat{\Theta}_c = \tau_c \Theta_c + (1 - \tau_c) \widehat{\Theta}_c$.

\begin{algorithm*}[!t]
\caption{The ANGC total discrepancy process under an environment for $E$ episodes (of maximum length $T$).}
\label{algo:angc_process}
\begin{algorithmic}
   \State {\bfseries Input:} environment $\mathbb{S}$, controller $\Theta_c$, generator $ \Theta_g$, deque memory $\mathcal{M}$, and constants $E$, $T$, $\alpha_e$, $\alpha_i$, $\epsilon_{decay}$, $\epsilon$, $\gamma$
   \Function{SimulateProcess}{$\mathbb{S}, E, T, \Theta_c, \Theta_g, \mathcal{M}, \alpha_e, \alpha_i, \epsilon, \epsilon_{decay}$}
        \State $r^i_{max} = 1$
        \For{$e = 1$ to $E$}
            \State $\mathbf{o}_t \leftarrow \mathbf{o}_0$ from $\mathbb{S}$ \Comment Get initial state/observation from environment
            \For{$t = 1$ to $T$}
              \LineComment Sample action $a_t$ according to an $\epsilon$-greedy policy
              \State $\mathbf{d}_t = \Call{Project}{\mathbf{o}_t, \Theta_c}$, $p \sim \mathbb{U}(0,1)$
              \State $\Big( p < \epsilon \rightarrow a_t \sim \mathbb{U}_d(1,A) \Big) \land \Big( p \ge \epsilon \rightarrow a_t = \arg\max_a \mathbf{d}_t \Big)$, \; $\mathbf{a}_t = \Call{toOneHot}{a_t}$
              \LineComment Get next state/observation from environment \& compute component reward signals
              \State $( r^e_t, \mathbf{o}_{t+1} ) \leftarrow \mathbb{S}(a_t)$, \; $(\Lambda, \mathcal{E}) = \Call{Infer}{[\mathbf{a}_t, \mathbf{o}_t], \mathbf{o}_{t+1}, \Theta_g}$
              \State $r^i_t = \sum_\ell ||\mathcal{E}[\ell]||^2_2$, \; $r^i_{max} = \max(r^i_t, r^i_{max})$, \; $r^i_t \leftarrow  \frac{r^i_t}{ r^i_{max} }$, \; $r_t = \alpha_e r^e_t + \alpha_i r^i_t$
              \LineComment Store transition and update weights from samples in memory
              \State Store $(\mathbf{o}_t, a_t, r_t, \mathbf{o}_{t+1})$ in $\mathcal{M}$
              \State $(\mathbf{o}_j, a_j, r_j, \mathbf{o}_{j+1}) \sim \mathcal{M}$ \Comment{Sample mini-batch of transitions from memory}
              \State $t_j = \begin{cases} 
                                      r_j & \mbox{if } \mathbf{o}_j \mbox{ is terminal} \\
                                      r_j + \gamma \max_a \Call{Project}{\mathbf{o}_{j+1}, \Theta_c} & \mbox{otherwise }
                                    \end{cases}$
              \State $\mathbf{a}_j = \Call{toOneHot}{a_j}$, \; $\mathbf{t}_j = t_j \mathbf{a}_j + (1 - \mathbf{a}_j) \otimes \Call{Project}{\mathbf{o}_j, \Theta_c}$
              \State $(\Lambda_c, \mathcal{E}_c) = \Call{Infer}{\mathbf{o}_j, \mathbf{t}_j, \Theta_c}$, \; $\Theta_c \leftarrow \Call{UpdateWeights}{\Lambda_c, \mathcal{E}_c, \Theta_c}$ \Comment Update controller $\Theta_c$
              \State $(\Lambda_g, \mathcal{E}_g) = \Call{Infer}{[\mathbf{a}_j, \mathbf{o}_j], \mathbf{o}_{j+1}, \Theta_g}$, $\Theta_g \leftarrow \Call{UpdateWeights}{\Lambda_g, \mathcal{E}_g, \Theta_g}$ \Comment Update generator $\Theta_g$
            \EndFor
            \State $\epsilon \leftarrow \max(0.05, \epsilon \cdot  \epsilon_{decay}) $
        \EndFor
   \EndFunction
\end{algorithmic}
\end{algorithm*}

\paragraph{The ANGC Agent:  Putting It All Together} 

At a high level, the ANGC operates, given observation $\mathbf{o}_t$, according to the following steps:
1) the NGC controller takes in $\mathbf{o}_t$ and uses it to produce a discrete action,
2) the ANGC agent next receives observation $\mathbf{o}_{t+1}$ from the environment, i.e., the result of its action,
3) the NGC generator runs the dynamics Equations \ref{eqn:ngc_predict}-\ref{eqn:ngc_correct} to find a set of hidden neural activity values that allow a mapping from $[\mathbf{a},\mathbf{o}_t]$ to $\mathbf{o}_{t+1}$ and then updates its own specific synaptic weights using Equations \ref{eqn:forward_update}-\ref{eqn:backward_update}, 
4) the reward $r_t$ is computed using the extrinsic/problem-specific reward plus the epistemic signal (produced by summing the layer-wise errors inside the generator, i.e., total discrepancy),
5) the NGC controller then runs the dynamics Equations \ref{eqn:ngc_predict}-\ref{eqn:ngc_correct} to find a set of hidden neural activity values that allow a mapping from $\mathbf{o}_t$ to $r_t$ and then updates its synaptic weights via Equations \ref{eqn:forward_update}-\ref{eqn:backward_update}, and, finally, 
6) the ANGC agent transitions to $\mathbf{o}_{t+1}$ and moves back to step 1.

The above step-by-step process shows that the NGC generator (forward model) drives information-seeking behavior (facilitating exploration better than that of random epsilon-greedy), allowing the ANGC agent to evaluate if an incoming state will allow for a significant reduction in uncertainty about the environment. The NGC controller (policy) is responsible for estimating future discounted rewards, balancing the seeking of a goal state (since the instrumental term represents the ``desire'' to solve the problem) with the search for states that will give it the most information about its environment. 
The controller keeps the agent focused on finding a goal state(s) and reinforcing discovered sequences of actions that lead to these goal states (exploitation) while the generator forces the agent to parsimoniously explore its world and seek elements that it knows least about but will likely help in finding goal states -- this reduces the number of episodes and/or sampled states needed to uncover useful policies.

Given that ANGC is inspired by active inference \cite{friston2011action}, the intuition behind our approach is that an agent reduces the divergence between its model of the world and the actual world by either:  
1) changing its internal model (the generator) so that it better aligns with sampled observations (which is why it seeks states with high epistemic/total discrepancy values), or 
2) changing its observations such that they align with its internal model through action, i.e., this is done through the controller tracking its problem-specific performance either through extrinsic reward values or other functions, i.e., prior preferences \cite{tschantz2020learning}. 
The ANGC agent's balancing act between finding goal states with better exploring its environment strongly relates to the rise of computational curiosity in the RL literature \cite{pathak2017curiosity} -- since we focus on using problem-specific rewards (instead of crafting prior preference distributions as in \cite{friston2009reinforcement,tschantz2020learning})
our agent’s instrumental term (in Equation \ref{eqn:reward}) is akin to extrinsic reward and our epistemic term is similar in spirit to intrinsic ``curiosity'' \cite{oudeyer2018computational,burda2018large} (error-based curiosity). Although our curiosity term is the total discrepancy of the NGC generator, this term connects ANGC to the curiosity-based mechanisms and exploration bonuses \cite{wu2016training} used to facilitate efficient extraction of robust goal-state seeking policies. Furthermore, the generator component of the ANGC agent connects our work with the recently growing interest in model-based RL where world models are integrated into the agent-environment interaction process, e.g., plan2explore \cite{sekar2020planning}, dreamer \cite{hafner2019dream}, etc. In a sense, one could view our ANGC as a simple, neurobiologically-plausible instantiation of these kinds of model-based RL approaches, offering a means to train similar systems without backprop. 
Many other bio-inspired algorithms, such as equilibrium propagation \cite{scellier2017equilibrium}, do not scale easily to RL problems due to expensive inference phases (whereas NGC's inference is faster -- see Appendix for details). 

In essence, the proposed ANGC framework prescribes the joint interaction of the controller and generator modules described above. 
At each time step, the agent, given observation $\mathbf{o}_t \in \mathcal{R}^{D \times 1}$ (which could contain continuous or discrete variables), is to perform a discrete action $a_t$\footnote{We focus on discrete actions in this study and leave generalization to continuous actions for future work.} and receive from its environment the result of its action, i.e., observation $\mathbf{o}_{t+1}$ and possibly an external reward signal $r^{ep}_t$. The controller is responsible for deciding which action to take next while the generator actively attempts to guess the (next) state of the agent's environment. Upon taking an action $a_t$, the generator's prediction is corrected using the values of the sensory sample drawn from the environment, allowing it to iteratively craft a compressed internal impression of the agent's world. The inability of the generator to accurately predict the incoming sensory sample $\mathbf{o}_{t+1}$ provides the agent with a strong guide to exploring its environment,  reducing its (long-term) surprisal and thus improving the controller's ability to extract an effective policy/plan.

The complete ANGC agent is specified in Algorithm \ref{algo:angc_process}\footnote{$\mathcal{E}[\ell]$ means ``retrieve the $\ell$th item in $\mathcal{E}''$.} and depicted in Figure \ref{fig:angc} (right). Note that Algorithm \ref{algo:angc_process} implements the full simulation of ANGC's inference and synaptic adjustment over an $E$-episode long stream (each episode is at most $T$ steps long -- where $T$ could vary with time). 
In addition to the target controller modification described earlier, we integrate experience replay memory $\mathcal{M}$ \cite{o2010play,mnih2015human} (implemented as a ring buffer with mini-batches sampled from stored transitions uniformly at random). This stabilizes the learning process by removing correlations in the observation sequence.

\section{Experiments}
\label{sec:experiments}

The performance of the ANGC agent is evaluated on three control problems commonly used in reinforcement learning (RL) and one simulation in robotic control. Specifically, we compare ANGC to a random agent (where actions are taken at each step uniformly at random), a deep Q-network (DQN) \cite{mnih2015human},
the intrinsic curiosity module (ICM) \cite{pathak2017curiosity},
and another powerful intrinsic curiosity baseline known as random network distillation (RnD) \cite{burda2018exploration}
on: 1) the inverted pendulum (cartpole) problem, 2) the mountain car problem 3) the lunar lander problem, and 4) the robot reaching problem. Details related to each control problem are provided in the appendix.


\paragraph{ANGC Agent Setup:}
\label{sec:angc_setup}
For all of the ANGC agents across all trials, we used fixed configurations of meta-parameters for each control task. We provide the key values chosen (based on preliminary experimentation) for the meta-parameters for all ANGC models in the appendix.

For all ANGC agents, $\alpha_e = \alpha_i = 1.0$ was used as the importance factors for both the epistemic and instrumental signals.
Both the controller and generative model were trained using a single, shared experience replay buffer with a maximum capacity of $N_{mem}$ transitions from which mini-batches of $N_{batch}$ transitions were sampled in order to compute parameter updates at any single time-step of each simulation. Each agent also uses an epsilon($\epsilon$)-greedy policy where $\epsilon$ was decayed at the end of each episode according to the rule:  $\epsilon \leftarrow \min(0.05, \epsilon * \epsilon_{decay})$ (starting $\epsilon=1$ at a trial's start). 

In addition, we experiment with an ablated form of our ANGC agent, i.e., Instr-ANGC, where the generator/forward model has been removed. This means that the Instr-ANGC only uses the extrinsic reward signal, allowing for a closer examination of what happens if the normal backprop-based neural model was just replaced with our NGC circuit.

\begin{figure*}
\centering
\begin{subfigure}[b]{.225\linewidth}
\includegraphics[width=\linewidth, frame]{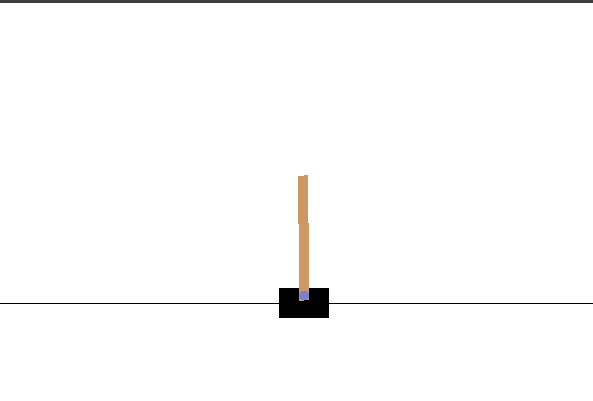}
\caption{Inverted pendulum.}\label{fig:cpole}
\end{subfigure}
\hspace{0.95cm}
\begin{subfigure}[b]{.225\linewidth}
\includegraphics[width=\linewidth, frame]{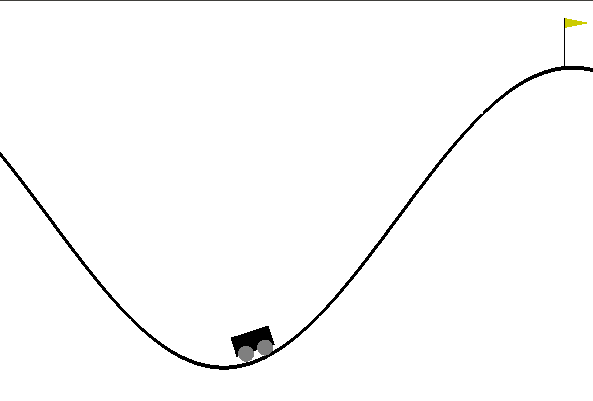}
\caption{Mountain car.}\label{fig:mcar}
\end{subfigure}
\hspace{0.95cm}
\begin{subfigure}[b]{.225\linewidth}
\includegraphics[width=\linewidth, frame]{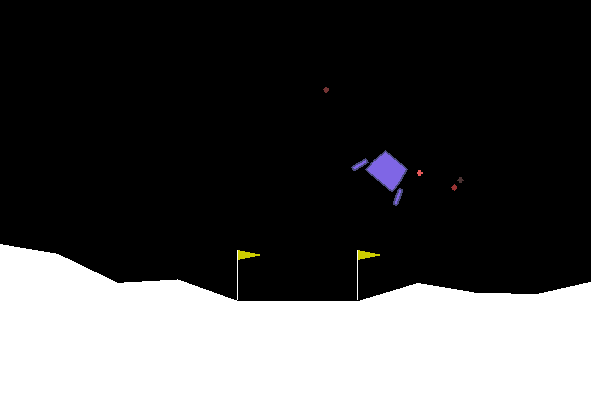}
\caption{Lunar lander.}\label{fig:lander}
\end{subfigure}
\begin{subfigure}[b]{.325\linewidth}
\includegraphics[width=\linewidth]{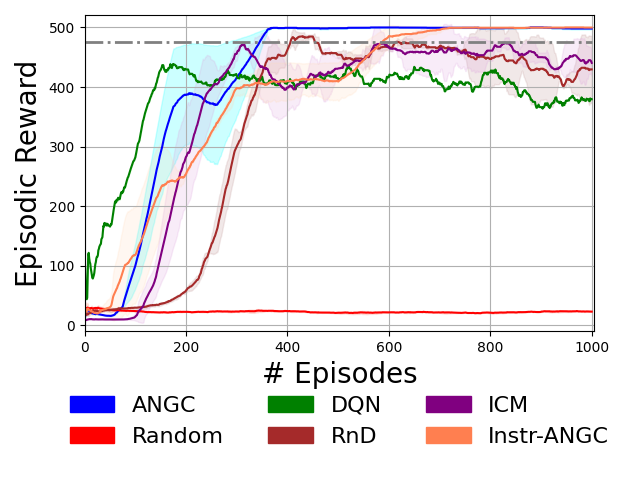}
\label{fig:cpole_mar}
\end{subfigure}
\begin{subfigure}[b]{.325\linewidth}
\includegraphics[width=\linewidth]{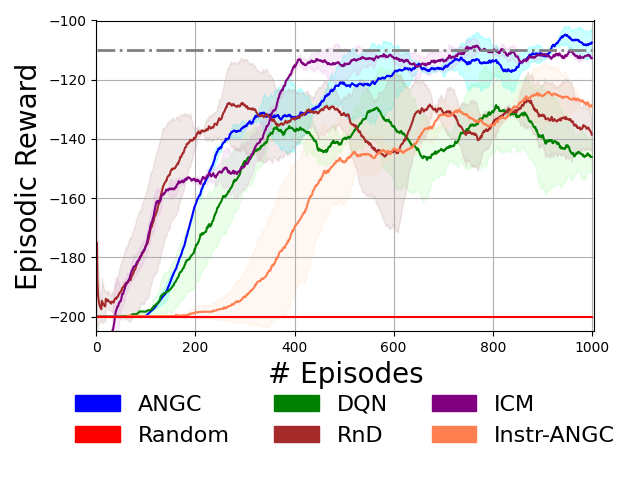}
\label{fig:mcar_mar}
\end{subfigure}
\begin{subfigure}[b]{.325\linewidth}
\includegraphics[width=\linewidth]{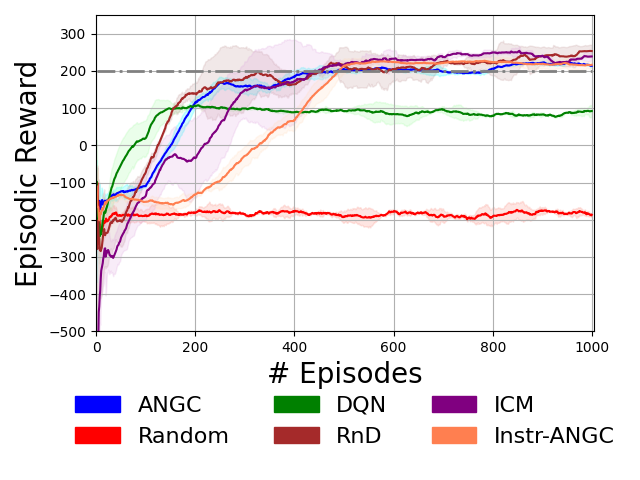}
\label{fig:lander_mar}
\end{subfigure}
\caption{Reward curves for ANGC and baselines (DQN, ICM, Rnd, and the ablated model Instr-ANGC). Mean and standard deviation over 10 trials are plotted. Dash-dotted, horizontal (gray) lines depict the problem solution threshold.}
\label{fig:results_mar}
\end{figure*}

\paragraph{Baseline Agent Setups:}
\label{sec:dqn_setup}
For the DQN, ICM, and RnD agents, we initially start with 90\% exploration and 10\% exploitation ($\epsilon=0.9$ ) and eventually begin decaying until the condition of 10\% exploration is reached, i.e., 90\% exploitation ($\epsilon=0.1$). The discount factor was tuned in the range of $\gamma = [0.91, 0.99]$. 
The linear rectifier was used as the activation function and Adam was used to update the weight values, except for ICM, where AdamW was found to yield more stable updates. For all models, each $W^\ell$ was initialized according to a centered Gaussian scaled by $\sqrt{2.0/(J_{\ell-1}+J_\ell)}$. The replay buffer size, the learning rate, the hidden dimensions, and number of layers were tuned -- hidden layer sizes were selected from within the range of $[32,512]$ and the number of layers was chosen from the set $[1,2,3]$. In the appendix, we provide best configurations used for each model.


\begin{figure*}[!t]
    \centering
    \begin{subfigure}[b]{.2\linewidth}
    \includegraphics[width=0.975\linewidth, frame]{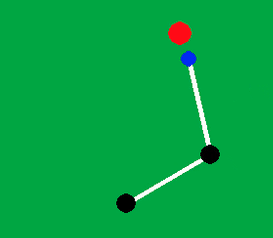}
    \label{fig:robot_arm}
    \end{subfigure}
    \begin{subfigure}[b]{.45\linewidth}
    \includegraphics[width=0.85\textwidth]{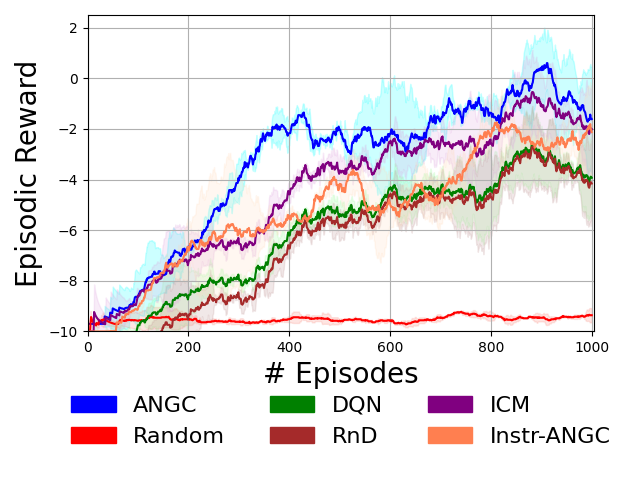}
    \label{fig:robot_results} 
    \end{subfigure}
    \caption{The robot arm reaching problem results. Mean and standard deviation over $10$ trials plotted.}
    \label{fig:robot_mar}
\end{figure*}

\paragraph{Results:}
\label{sec:results}
In Figures \ref{fig:results_mar} and \ref{fig:robot_mar}, we visualize the accumulated reward as a function of the episode count (over the first $1000$ episodes, see Appendix for expanded results), smoothing out the curves by plotting the moving average reward as a function of the episode count, i.e., 
$\mu_t = 0.1 r_t + 0.9 \mu_{t-1} $ 
. Results are averaged over $10$ trials and the plots present both the mean (darker color central curve) and standard deviation (lighter color envelope). In each plot, a dash-dotted horizontal line depicts the threshold for fully solving each task.

It is immediately apparent from our reward plots, across all four benchmarks, that the ANGC agent is notably competitive with backprop-based, intrinsic curiosity models (ICM, RnD), extracting a better policy than models that do not incorporate intelligent exploration (the DQN). 
This highlights the value of our ANGC framework for designing agents -- for each of the benchmarks we investigate, fewer episodes are required by ANGC agents to generalize well and even ultimately solve a given control problem (as indicated by their ability to reach each problem's solution threshold).
Furthermore, the ANGC offers stable performance in general, where models like the RnD sometimes do not (as on mountain car problem), and is notably quite competitive with ICM in general, even outperforming it on the more complex robotic arm control task.
We also note that the DQN reaches its goal quickly in some cases but struggles to maintain itself at the solution threshold, fluctuating around that range (requiring more episodes to fully stabilize).

Crucially, observe that the ANGC agent is capable of effectively tackling control problems involving extremely sparse rewards (or nearly non-existent reward signals) as indicated by its early strong performance on the mountain car and robot reaching problems, which are arguably the hardest of the problems examined in this study. The ANGC's effectiveness on these problems is, we argue, largely due to its own internally generated epistemic modulation factor $r^{ep}_t$ (this is empirically corroborated by the Instr-ANGC's worse performance than the full ANGC). In other words, the ANGC agent explores states that surprise it most, meaning it is most ``curious'' about states that yield the highest magnitude total discrepancy (or the greatest free energy). This feature presents a clean NGC implementation of the epistemic term key to the active inference framework \cite{friston2017active_curosity} which, theoretically, is meant to encourage a more principled, efficient form of environmental exploration. 
Furthermore, this term, much akin to intrinsic curiosity models \cite{pathak2017curiosity}, would allow the agent to operate in settings where even no external reward is available.

Note that the intent of this study was not to engage in a performance contest given that there is a vast array of problem-tuned, neural-based RL approaches that attain state-of-the-art performance on many control tasks. Instead, the intent was to present a promising alternative to backprop-based approaches and demonstrate that ANGC can acquire good policies on control problems that DQN-based models can.
While our ANGC agent results are promising, integrating additional mechanisms typically used in deep RL would be a fruitful next step.
Since our framework has proven to be compatible with commonly-used RL heuristics such as experience replay and target network stability, integrating other heuristics would help to further improve performance. In the appendix, we discuss the limitations of the ANGC framework, the ANGC framework's relationship with free energy optimization, examine ANGC in the context of related work in RL and planning as inference, and conduct further analysis on the control problems presented above.

\section{Conclusion}
\label{sec:conclusion} 
In this paper, we proposed active neural generative coding (ANGC), a framework for learning goal-directed agents without backpropagation of errors (backprop). We demonstrated, on four control problems, the effectiveness of our framework for learning systems that are competitive with backprop-based ones such as the deep Q-network and powerful intrinsic curiosity variants.
Notably, our framework demonstrates the value of leveraging the neuro-biologically grounded learning and inference mechanisms of neural generative coding to dynamically adapt a generative model that provides intrinsic signals (based on total discrepancy) to augment problem-specific extrinsic rewards.
Furthermore, given its robustness, the ANGC framework could prove useful in more complex environments (e.g., the Atari games, challenging robotic problems), offering an important means of implementing longer-term planning-as-inference.

{
\small
\bibliographystyle{IEEEtran}
\bibliography{ref}
}

\appendix

\input{supplement_arxiv.tex}

\end{document}

%% file: supplement_arxiv.tex
\section*{Appendix}

\section{Additional Model and Algorithmic Details}
\paragraph{Algorithm CC - The NGC Circuit:}
In the main paper, the NGC section described the neural dynamics required for figuring out the values of an NGC network given an input and output as well as the weight update rules (which, for any matrix, is simply the product of the error neurons at the current layer and the actual post-activation values of the layer above it). To tie together these details further, in this section of the appendix, we present the full pseudocode for simulating a complete NGC circuit in Algorithm CC \ref{algo:ngc_process}.
Note that a full ANGC agent utilizes two NGC circuits -- one for implementing the controller/policy model and one for implementing the generator/forward model.

\begin{algorithm*}[!t]
\floatname{algorithm}{Algorithm CC}
\caption{The NGC model's projection, inference, and weight update routines.}
\label{algo:ngc_process}
\begin{algorithmic}
   \State {\bfseries Input:} Sample $(\mathbf{x}^i,\mathbf{x}^o)$ (from sensory stream), $\beta$, $\beta_e$, $\gamma_v$, $\gamma_e$, $\eta$, $K$, $\Theta$, choice of $\phi^\ell$ \& $g_\ell$, and $c_\epsilon = 10^{-6}$
   \Function{Project}{$\mathbf{x}^i, \Theta$}
        \LineComment Ancestrally project pattern through model
        \State $\mathbf{\bar{z}}^L = \mathbf{x}^i$
        \For{$\ell = L-1$ to $0$}
            \State Get $\mathbf{W}^{\ell+1}$ from $\Theta$, \; $\mathbf{\bar{z}}^\ell =  g_\ell(\mathbf{W}^{\ell+1} \ast \phi^{\ell+1}(\mathbf{\bar{z}}^{\ell+1}) )$
        \EndFor
        \State \textbf{Return} $\mathbf{\bar{z}}^0$
   \EndFunction
   \Function{Infer}{$\mathbf{x}^i, \mathbf{x}^o, \Theta$}
        \LineComment Simulate stimulus presentation time over $K$ steps
   		\State $\mathbf{z}^0 = \mathbf{x}^o$, $\mathbf{z}^1 = \mathbf{0}, \cdots, \mathbf{z}^\ell = \mathbf{0}, \cdots, \mathbf{z}^L = \mathbf{x}^i$
   		\State $\mathbf{e}^0 = \mathbf{z}^0 - 0, \mathbf{e}^1 = \mathbf{0}, \cdots, \mathbf{e}^\ell = \mathbf{0}, \cdots, \mathbf{e}^L = \mathbf{0}$
   		\For{$k = 1$ to $K$}
   		\LineComment Correct latent states given current value of error units
   		\For{$\ell = 1$ to $L$}
   		    \State Get $\mathbf{E}^\ell$ from $\Theta$, \; $\mathbf{z}^\ell \leftarrow \mathbf{z}^\ell + \beta ( -\gamma_v \mathbf{z}^\ell  -\mathbf{e}^\ell + ( \mathbf{E}^\ell \ast \mathbf{e}^{\ell - 1})  + \vartheta(\mathbf{z}^\ell) ) )$ \Comment Maps to Equation 1 
   		\EndFor
   		\LineComment Compute layer-wise predictions and error neuron values
   		\For{$\ell = L-1$ to $0$}
   		    \State Get $\mathbf{W}^{\ell+1}$ from $\Theta$, \; $\mathbf{\bar{z}}^\ell =  g_\ell(\mathbf{W}^{\ell+1} \ast \phi^{\ell+1}(\mathbf{z}^{\ell+1}) )$, \; $\mathbf{e}^\ell = \frac{1}{2\beta_e} (\phi^\ell( \mathbf{z}^\ell ) - \mathbf{\bar{z}}^\ell)$ \Comment Equation. 2 
   		\EndFor
   		\EndFor
   		\State $\Lambda = \{\mathbf{z}^0, \mathbf{z}^1, \cdots, \mathbf{z}^\ell, \cdots, \mathbf{z}^L\}$, $\mathcal{E} = \{\mathbf{e}^0, \mathbf{e}^1, \cdots, \mathbf{e}^\ell, \cdots, \mathbf{e}^{L-1}\}$
        \State \textbf{Return} $\Lambda, \mathcal{E}$
    \EndFunction
    \Function{UpdateWeights}{$\Lambda, \mathcal{E}, \Theta$}
    \LineComment Adjust synaptic weight adjustments given states and error neurons
    \For{$\ell = 1$ to $L$}
        \State Retrieve $(\mathbf{W}^\ell, \mathbf{E}^\ell)$ from $\Theta$
        \State $\Delta \mathbf{W}^\ell = \mathbf{e}^\ell \ast (\phi^{\ell+1}(\mathbf{z}^\ell))^T$, \; $\Delta \mathbf{W}^\ell \leftarrow  \frac{\Delta \mathbf{W}^\ell}{||\Delta \mathbf{W}^\ell||_2 + c_\epsilon}$, \;
        $\Delta \mathbf{W}^\ell \leftarrow \Delta \mathbf{W}^\ell \otimes \mathbf{M}^\ell_W$
        \State $\Delta \mathbf{E}^\ell = \gamma_e (\Delta \mathbf{W}^\ell)^T$, \; $\Delta \mathbf{E}^\ell \leftarrow \frac{\Delta \mathbf{E}^\ell}{||\Delta \mathbf{E}^\ell||_2 + c_\epsilon}$, \;
        $\Delta \mathbf{E}^\ell \leftarrow \Delta \mathbf{E}^\ell \otimes \mathbf{M}^\ell_E$
        \State $\mathbf{W}^\ell \leftarrow \mathbf{W}^\ell + \eta \Delta \mathbf{W}^\ell$, \; $\mathbf{E}^\ell \leftarrow \mathbf{E}^\ell + \eta \Delta \mathbf{E}^\ell$ \Comment Can alternatively use Adam or RMSprop
        \State $\mathbf{W}^\ell \leftarrow \frac{2\mathbf{W}^\ell}{||\mathbf{W}^\ell||_2 + c_\epsilon}$, \; $\mathbf{E}^\ell \leftarrow \frac{2\mathbf{E}^\ell}{||\mathbf{E}^\ell||_2 + c_\epsilon}$ \Comment Override current values of $(\mathbf{W}^\ell, \mathbf{E}^\ell)$ in $\Theta$
    \EndFor
    \State $\widehat{\Theta} = \{\mathbf{W}^0,\mathbf{E}^0,\mathbf{W}^1,\mathbf{E}^1,\cdots,\mathbf{W}^L,\mathbf{E}^{L}\}$
    \State \textbf{Return} $\widehat{\Theta}$
    \EndFunction
\end{algorithmic}
\end{algorithm*}

\paragraph{Explanation of the NGC Controller:}
Informally, the controller (or policy model) is ultimately meant to: 
\begin{itemize}
    \item choose from a set of discrete actions such that it chooses the action that yields the greatest extrinsic (problem-specific) reward and also leads to the next state of the environment that the agent stands to gain the most information from -- ideally, after so many episodes, the agent should find that most aspects of its environment relevant to reaching a goal state are not that surprising (uncertainty/total discrepancy over those states would be low), and
    \item learn how to better estimate the future, discounted reward values for various actions that it could take at a given state of the environment (the combined extrinsic and intrinsic/uncertain-reducing signals)
\end{itemize}
In essence, the controller tries to predict the reward values associated with each action and then uses a reward signal, created through the combination of both instrumental and epistemic signals, plus the weighted (intrinsic) term contributed by the target NGC network (this is very similar to the process of TD(0) or Q-learning). 
In sum, what the algorithm provided in the main paper (see section titled ``Generalizing to Active Neural Coding'') is depicting is that, at any time step, given an observation vector $\mathbf{o}_t$ and the next transition $\mathbf{o}_{t+1}$, the agent proceeds as follows (if we ignore the replay buffer and target network extensions):
\begin{enumerate}
    \item the NGC controller (policy model) is used to pick an action and the agent obtains $\mathbf{o}_{t+1}$,
    \item the NGC generator (forward model) attempts to predict $\mathbf{o}_{t+1}$ by running the dynamics equations (Equations 1-2) of the NGC section (and updates its weights once it has found a set of hidden activities),
    \item the (dynamically normalized) reward $r_t$ is computed by combining the total discrepancy of the generator ($r^{ep}_t$) with the task's extrinsic reward ($r^{in}_t$), and
    \item the NGC controller runs the dynamics equations (Equations 1-2) in the NGC section to find hidden layer values that would satisfy the mapping between state/observation $\mathbf{o}_t$ to $r_t$ (and updates its weights once it has found a set of hidden activities).
\end{enumerate}

\section{Active Inference and ANGC}

\paragraph{Goal-Directed Behavior:}
In this paper, we use phrase ``goal-directed'' to refer to the case that an intelligent agent has a goal to reach as opposed to pure exploration/exploration-only problems -- even in Markov decision problems/processes, there are usually goals or goal states to reach. Active inference often distinguishes between two key objectives for a given agent: an instrumental (extrinsic / pragmatic / goal-directed) objective and an epistemic (intrinsic / uncertainty-reducing / information-seeking) objective \cite{tschantz2020learning}. In effect, ANGC was designed to follow the same overall form -- its signal $r^{in}_t$ serves as the instrumental objective and its its signal $r^{ep}_t$ serves as the epistemic objective.

\paragraph{Relationship to Free Energy Minimization:}

In ANGC, it is important to understand that, even though the system is formulated in the main paper such that the controller is maximizing long-term discounted reward, free energy \cite{friston2010free} is still being minimized as a result of the NGC generator reducing its total discrepancy (when predicting future states). 
Specifically, if one labels the ANGC's generator as the ``transition model''\footnote{Here, we have assumed that our agent’s encoder and decoder are identity matrices which means that we have collapsed the observation space to be equal to a latent state space. Note that by removing this assumption, as we discuss later in this appendix, ANGC could be scaled up to handle pixel-valued inputs (particularly if the encoder/decoder are convolutional and utilize weight sharing).} and the controller as the ``expected free energy estimation model'', which aligns ANGC with terminology used in work on deep active inference \cite{millidge2020deep,ueltzhoffer2018deep}), then an ANGC agent, much in the core spirit of active inference, can be understand as seeking out environment states (observations) that are most informative, i.e., those states that allow for the greatest reduction in uncertainty (since the agent is most ``surprised'' when the epistemic value is high).
As a result, (approximate) free energy, of which total discrepancy is a special instance of \cite{ororbia2020neural}), will still be minimized because, upon selecting a state and transitioning, the agent's transition model will update its synaptic parameters to ensure that it is able to better predict that state later in the future. In other words, the next time that an environment state is encountered again, the agent's measured discrepancy for that particular state (or a very similar state) will be lower. The maximization of the epistemic reward is meant to drive the ANGC agent towards states that will be best for it to discover early on when developing good policies as it reduces uncertainty about its environment.


\paragraph{Prior Preferences Beyond a Reward Scalar: }
If rewards are viewed as log priors, i.e., $p(\mathbf{o}) \propto \exp(r(\mathbf{o}))$ \cite{friston2012active}, then other choices of  $p(\mathbf{o})$ are possible. This generality is afforded by the complete class theorem \cite{brown1981complete,wald1947essentially,daunizeau2010observing}, which says that, for any pair of reward functions/preferences and choice behavior, there exists a prior belief that render the choices Bayes optimal. Understanding the theoretical backing of the complete class theorem means that our ANGC framework, despite the fact that it commits to a particular form of neural processing (to provide a concrete implementation for simulation), could be written down in terms of Bayesian decision processes even though the form we present does not explicitly do so.

There are multiple alternatives to reward-based prior preferences. For example, in the mountain car problem, one could design a prior function to replace the problem's standard reward function, i.e., a Gaussian distribution centered over the coordinates of the flag post (the mean is hard-coded to this coordinate) with a standard deviation of $\sigma = 0.1$, much as in \cite{friston2009reinforcement}. Alternatively, one utilize another neural model to serve as a dynamic prior -- this model would be trained to learn such a prior by imitating expert data \cite{shin2021prior}. The ANGC framework is readily amenable and capable of operating with prior preferences such as those just described.

\section{Discussion and Related Work}
\label{sec:discussion}

\subsection{Limitations of ANGC:}
\label{sec:limitations}
Despite the promise that the proposed ANGC agent framework offers, the current implementation/design we propose does come with several drawbacks. First, if one does not design an NGC/ANGC model with capacity in mind at the beginning, i.e., considering the total number of weights initialized when choosing number of layers and layer dimensionalities, it is very easy to create agents that require a great deal of storage memory. An ANGC agent operates with both forward generative weights \textbf{and} error correction weights, meaning that increasing the size of any single layer by adding extra processing elements can result in significant increases in required memory (as doing so increases not only the number of forward synapses but also the number of error synapses). 
Second, while we have demonstrated that ANGC agents acquire good policies reasonably early (reducing the amount of data needed when learning online from a dynamic environment) compared to several backprop-based reinforcement learning (RL) models, one might be wary of the fact that, per time step/input pattern presented to the model, additional computation is required to search for latent neural activities that minimize an ANGC agent's total discrepancy (i.e., iterative inference is naturally more expensive than a single feedfoward pass). Even though, in this paper, our ANGC/NGC implementation only required a relatively small number of steps (we found values $K=10$ and $K=20$ to be sufficient) for the problems we examined, for more complex patterns of a higher dimensionality, it is likely that a higher value of $K$ will be required to settle to useful neural activities before updating the agent's synaptic weights. To combat this issue, one could consider designing algorithms for amortized inference, as has been done to speed (to varying degrees) up neural models, e.g., predictive sparse decomposition \cite{kavukcuoglu2010fast}. Another option is to design software and hardware that fully exploits the layer-wise parallelism inherent to the NGC/ANGC's inference/learning computations (since the method is not forward and backwards-locked, i.e., each layer of neurons and the weights that immediately connect to them could be placed on their own dedicated GPU if a multi-GPU setup was available).

While the problems we examined varied in difficulty and certainly prove challenging for most RL models\footnote{The mountain car, despite its low dimensionality and simplicity, is, in fact, a very challenging problem even for modern-day RL systems given that its reward function is extremely sparse and requires intelligent exploration to properly and effectively solve.}, we were not unable (due to limited available computational resources) to examine the ideal setting with which we believe that the ANGC framework will yield the most interesting and important advances in RL research -- high-dimensional, imaged-based problems. In its current form, the ANGC framework would not scale well without first designing a few additional components that address the following issues:
\begin{enumerate}
    \item the generative model currently operates by taking in as input $\mathbf{o}_t$ to then predict $\mathbf{o}_{t+1}$, which means that as the dimensionality of the observation space increases (as would happen with large image pixel feeds) both the input and output size of the generative model would also increase (these dimensions would, unfortunately, scale linearly with respect to the dimensionality). To circumvent this, our ANGC's generator/transition model could be extended leverage a jointly-learned encoder and decoder to map from a lower-dimensional latent space to a high-dimensional pixel observation/state space (this is the subject of our future follow-up work),
    and,
    \item since processing natural images is challenging and often requires going beyond fully-connected, dense neural structures -- the ANGC would very likely need to be generalized to the case of locally-connected weight structures and convolution operators.
    Work in this direction \cite{ororbia2020large} offers some potentially useful local update rules for convolution but consideration on how to efficiently integrate convolution will be key (as well as methodology for quickly deciding a reasonable number of filters for acquiring useful feature detectors, which is, in of itself, a challenging part of tuning convolutional neural networks).
\end{enumerate}
We would like to point out that, although it is true that our current version of ANGC would face difficulty scaling up to very high-dimensional observation spaces,
our method would scale/operate well in reasonably-sized spaces, e.g., $32 \times 32$ or $64 \times 64$ pixel images would be reasonable given a working GPU, as in \cite{tschantz2020learning}).

Despite NGC's more expensive per-time-step calculation cost (and increased memory usage) as compared to backprop, the computational gain from using NGC over backprop is generally in terms of sample efficiency -- related work on NGC has demonstrated that fewer samples are needed to construct a good generative model as compared to backprop \cite{ororbia2020neural}. 
This improved sample efficiency balances out the increase in per-input computation \cite{marino2018iterative,ororbia2020continual,ororbia2020neural}. Nevertheless, we emphasize that the cost of an NGC model can be significantly reduced by exploiting tensor operations that leverage its internal parallelism and (often) sparse activities as well as introducing a procedure to amortize the cost of its iterative inference. 

\subsection{Neurocognitive Issues, Merits, and Related Work}

First and foremost, the fundamental building blocks of our ANGC agent align with the plethora of predictive processing computational models that have been proposed to explain brain function \cite{rao1999predictive,friston2005theory,bastos2012canonical}. This provides a desirable grounding of our model in computational cognitive neuroscience by connecting it to a prominent Bayesian brain theory as well as with established general principles of neurophysiology (for example, it combines diffuse inhibitory feedback connections with driving feedforward excitatory connections). In addition, the synaptic weight adjustments computed in our framework are local, corresponding to, if we include the factor $1/2\beta_e$ (and ideally replace it with the learnable precision matrix of \cite{ororbia2020neural}, a three-factor (error) Hebbian update rule. While there are many elements of our NGC building block that preclude it from serving as a complete and proper computational model of actual neural circuitry, e.g., synapses are currently allowed to be negative and positive, neurons are communicating with real-values instead of spikes \cite{wacongne2012neuronal}, etc., it represents a step forward towards a computational framework that facilitates a scalable simulation of neuro-biologically plausible computation that generalizes well on complicated statistical learning problems. We also note that our implementation further makes several design choices for the sake of computational speed and therefore only represents one possible implementation of predictive processing generalized for action-driven adaptation to dynamic environments. However, despite these limitations, our ANGC framework offers a promising path towards viable neural agents that learn without backprop, no longer subject to the algorithm's particular constraints  \cite{bengio2015towards}. Practically, one could leverage the parallelism afforded by high performance computing to potentially simulate very large, non-differentiable neural circuits given that our NGC modules offer natural layer-wise parallel calculations and do not suffer from the forward and backward-locking problems that plague backprop \cite{jaderberg2017decoupled}.
Furthermore, our ANGC contributes to the effort to create more biologically-motivated models and update rules either based on or for reinforcement learning \cite{mazzoni1991more,alexander2018frontal,yamakawa2020attentional} and motor control.

Second, key to our overall ANGC agent design is the notion of novelty or surprise \cite{barto2013novelty}, which is what provides our neural system with a means to explore an environment beyond a uniform random sampling scheme. This connects our ANGC framework to the family of RL models that have been designed to generate intrinsic reward signals \cite{schmidhuber1991possibility,storck1995reinforcement,singh2004intrinsically,oudeyer2009intrinsic}, which are inspired by the psychological concept of curiosity inherent to human agents \cite{ryan2000intrinsic,silvia2012curiosity}. Crucially, curiosity provides an agent with the means to acquire new skills that might prove useful for maximizing rewards in downstream tasks -- as a result, one could view the total discrepancy signal produced by the ANGC's generator (to create $r^{ep}_t$) as one potential source of ``curiosity'' that drives the agent, even in the presence of nearly no external reward signals (as was the case in our mountain car experiment).
Note that there are many other forms of intrinsic reward signals such as those based on policy-entropy \cite{rawlik2013probabilistic,haarnoja2018soft}, information gain \cite{houthooft2016vime,shyam2019model}, prediction error \cite{pathak2017curiosity,burda2018exploration},
state entropy \cite{lee2019efficient}, state uncertainty \cite{o2018uncertainty}, and empowerment \cite{leibfried2019unified,mohamed2015variational}. 
Since our agent employs and adapts a dynamic generative model to produce the necessary signals to drive its curiosity, our agent also contributes to the work in improving model-based RL through the use of world models \cite{ha2018recurrent,nagabandi2018neural,chua2018deep,hafner2019learning}. However, world models in modern-day, model-based RL are learned with backprop whereas our ANGC agent uses the same parallel neural processing and gradient-free weight updating as the controller, further obviating the need for learning the entire system in pieces or using evolution to learn an action model \cite{ha2018recurrent}.
Notably, generative world models could be useful when integrated into powerful planning algorithms \cite{rubinstein1997optimization,williams2016aggressive}. Utilizing our ANGC's generator for long-term planning will be the subject of future work.

Finally, our ANGC agent framework offers a simple predictive processing interpretation of active inference and the more general, theoretical planning-as-inference.
Planning-as-inference (PAI) \cite{botvinick2012planning} broadly encapsulates the view that a decision-making agent utilizes an internal cognitive model that represents the agent's future as the joint probability distribution over actions, (outcome) states, and rewards. 
Active inference \cite{friston2017active_curosity}, which is a particular instantiation of PAI, suggests that agents select actions such that they maximise the evidence for an internal model that is also biased towards the agent’s preferences.
In Equation 3,
we took advantage of the complete class theorem \cite{brown1981complete} to develop a simple weighted sum of the two signals central to the general active inference optimization objective -- an instrumental term (which drives the agent towards a goal or preferred state of the world) and an epistemic term (which drives the agent to search through its environment by attending to states that surprise it the most). While we focus on scalar signals, which would likely be components of the error function that is embodied in the firing rates of dopamine neurons \cite{montague1994predictive,montague1996framework,schultz1997neural}, one could instead use encodings of goal states, priors, or other functions that could facilitate more complex behavior and longer-term planning in an ANGC agent.
Neural process theory has also been developed for active inference \cite{friston2009predictive,parr2017activeworld} which could be used to further modify our framework towards greater neurobiological plausibility.

\paragraph{Neurobiologically-Inspired RL and Learning:} 
Hebbian RL \cite{najarro2020meta} is a meta-learning RL scheme that relies on an evolutionary strategy to adapt its key model parameters (such as the Hebbian coefficients). In contrast, the ANGC framework focuses on crafting a local, neurobiological scheme that does not depend on evolution.
Furthermore, \cite{najarro2020meta} focused on the challenge of continual RL, whereas this study was focused on demonstrating the viability of a backprop-free framework for handling RL problems. Nevertheless, it would be useful to investigate how ANGC would handle continual RL and how aspects of Hebbian RL could be integrated into this computational framework. Interestingly enough, the Hebbian update rule within the Hebbian RL framework, we speculate, could complement ANGC given that related work has shown Hebbian co-occurrence rules offer useful forms of regularization \cite{o1998six,ororbia2020continual}.
In other bio-inspired approaches, the learning process is decomposed based on the source of reward, which serves as a goal for any given subtask \cite{zhou2004biologically}. In addition, an artificial emotion indication (AEI) is assigned for each subtask, where the AEI is responsible for predicting the reward component associated with a given subtask.
Another approach demonstrated that the backprop algorithm created issues for the learning process in modern day neural models, particularly for those for RL models meant to obtain desired solutions in minimal time and proposed a bio-inspired alternative to train RL systems \cite{weibio-inspired}. Specifically in this work, the network weights undergo spontaneous fluctuations and a reward signal is responsible for modulating the center and amplitude of these fluctuations . This ensures faster and enhanced convergence such that the RL network model acquires desired behavior.

While various biologically-inspired algorithms have been proposed to replace backprop \cite{movellan1991contrastive,o1996biologically,lee2015difference,lillicrap2016random,scellier2017equilibrium,guerguiev2017towards,whittington2017approximation,ororbia2018biologically,ororbia2020large} over the past several years, research has largely focused on developing in the context of classification with some exceptions in generative modeling and in processing temporal data \cite{wiseman2017training,ororbia2020continual,manchev2020target,ororbia2020neural}. Although any backprop-alternative could be, in principle, reformulated for reinforcement learning in dynamic environments, many such algorithms either require very long chains of computation to simulate their inference, such as contrastive Hebbian learning \cite{movellan1991contrastive} and equilibrium propagation \cite{scellier2017equilibrium}, or do not offer complete approaches to resolving all biological criticisms of backprop, for example, representation alignment \cite{ororbia2018biologically}, feedback alignment \cite{lillicrap2016random}, and target propagation \cite{lee2015difference} algorithms do not resolve the update and backwards locking problems \cite{jaderberg2017decoupled} while approaches such as \cite{whittington2017approximation} do not resolve the weight transport problem and require activation function derivatives. In contrast, NGC has been shown to offer one viable manner in which to resolve many of backprop's key criticisms and does not require multiple expensive inference phases \cite{ororbia2020neural} and has already been empirically shown to work well in developing temporal generative models without unfolding the computation graph through time (as is critical in backprop through time) \cite{ororbia2020continual,ororbia2020neural}. Since the motivation behind this work is to develop a flexible agent that embodies active inference yet conducts inference and synaptic adjustment in a neurocognitively-plausible fashion, NGC serves as a strong candidate computational framework and starting point.

\subsection{Potential Societal Impact} 
\label{sec:broader_impact}

The framework proposed in this paper presents an alternative to backprop for (online) training of ANN-based agents in the reinforcement learning setting, specifically offering robustness to scenarios where the external reward is sparse. The design of our approach, which obviates the need for expensive calculations (such as those related to point-wise derivatives) and offers a great deal of potential parallelism (especially since each layer conducts its operations independently of the others during generation), offers a path towards reducing the carbon footprint of ANN training by using fewer computational resources. However, if the parallelism of NGC/ANGC is not exploited and a strictly sequential implementation is used, the carbon footprint would be increased since additional computation is required at each step in time when processing observations (despite the fact the ANGC converges to useful policies sooner) and that more memory is required to store error synapses.
Since our approach is strongly inspired by human brain dynamics, our algorithm might have broad reach across the statistical learning and computational neuroscience communities in the effort to build biologically-motivated models/algorithms that generalize well. 
This research is a step towards building ANNs that learn and behave a bit more like humans, opening up a computational pathway that bridges our understanding of human intelligence and artificial neural systems.

The potential negative societal impact of the proposed ANGC computational framework is indirect -- while the model and algorithmic framework we develop is foundational in nature, it could potentially affect the myriad of applications and systems currently in development today or to be developed in the future. As a result, at best, the same negative consequences that result from using and integrating backprop-based ANNs are still present when using NGC as the training process/procedure instead. At worst, given that we have presented promising results on control problems with sparse rewards, the ANGC framework could facilitate the development of potentially better-performing robotic agents (given the robotic environments are not only complex but often contain sparse rewards at best) that might be used in military applications that might result in the loss of life, such as drones. Despite the benefits that ANGC offers to the machine learning and cognitive neuroscience communities, one should consider the drawbacks of building powerful neural systems to drive applications.


\section{Additional Experimental Details}

\begin{table*}[!t]
\begin{center}
\begin{tabular}{l|ll|ll|ll}
 & \multicolumn{2}{c}{\textbf{Cartpole}} & \multicolumn{2}{|c}{\textbf{Mountain Car}} & \multicolumn{2}{|c}{\textbf{Lunar Lander}}\\
 & \textbf{Controller} & \textbf{Generator} & \textbf{Controller} & \textbf{Generator} & \textbf{Controller} & \textbf{Generator}\\
 \hline 
 $\phi(\cdot)$ & ReLU  &  ReLU & ReLU6  &  ReLU6 & ReLU6  &  ReLU6 \\
 Dims & $[256, 128]$ & $[256, 128]$  & $[128,128]$ & $[128,128]$ & $[512,256]$ & $[128,128]$ \\
 Rule & RMSprop & Adam & Adam & Adam & Adam & Adam   \\
 $\eta$ & $0.0005$ & $0.001$ & $0.001$ & $0.001$ & $0.001$ & $0.001$\\
 $\epsilon_{decay}$ & $0.97$ & -- & $0.95$  &  -- & $0.995$  &  -- \\
 $C$ & $100$ & -- & $200$  &  -- & $200$  &  -- \\
 $\gamma$ & $0.99$ & -- & $0.99$  &  -- & $0.99$  &  -- \\
 $N_{mem}$ & $10^6$ & $10^6$ & $500,000$ & $500,000$ & $500,000$ & $500,000$ \\
 $N_{batch}$ & $256$ & $256$ & $128$ & $128$ & $256$ & $256$ \\
 \hline
\end{tabular}
\caption{For each control task, below are meta-parameter configurations used for the ANGC agents.}
\label{table:configs_angc}
\begin{tabular}{l|lll|lll|lll}
 & \multicolumn{3}{c}{\textbf{Cartpole}} & \multicolumn{3}{|c}{\textbf{Mountain Car}} & \multicolumn{3}{|c}{\textbf{Lunar Lander}}\\
 & \textbf{ICM} & \textbf{RnD} & \textbf{DQN} & \textbf{ICM} & \textbf{RnD} & \textbf{DQN} & \textbf{ICM} & \textbf{RnD} & \textbf{DQN} \\
 \hline 
 $\phi(\cdot)$ & ReLU  &  ReLU &  ReLU&  ReLU&  ReLU & ReLU  &  ReLU & ReLU  &  ReLU \\
 Dims & $[256, 256]$ & $[256, 128]$ & $[256,256]$ & $[256,128]$ & $[256,256]$  & $[256,192]$ & $[512,256]$ & $[512,256]$  & $[512,384]$ \\
 Rule & AdamW & Adam & Adam & Adam & Adam & Adam & AdamW & Adam & Adam   \\
 $\eta$ & $0.0001$ & $0.0003$  & $0.0005$ & $0.0004$ & $0.0005$  & $0.0002$ & $0.0005$ & $0.0001$  & $0.0002$ \\
 $\epsilon_{decay}$ & $0.97$ & $0.98$  & $0.95$ & $0.95$  &  $0.96$  & $0.95$ & $0.990$  &  $0.993$  & $0.990$ \\
 $C$ & $128$ & $128$  & $128$ & $256$  &  $256$  & $256$ & $256$  &  $256$  & $256$\\
 $\gamma$ & $0.992$ & $0.991$  & $0.990$ & $0.992$  &  $0.992$  & $0.992$ & $0.993$  &  $0.992$  & $0.991$ \\
 $N_{mem}$ & $10^6$ & $10^6$  & $10^6$ & $5^5$ & $5^5$ & $6^5$ & $5^5$ & $5^5$  & $6^5$ \\
 $N_{batch}$ & $256$ & $256$  & $256$ & $256$ & $256$  & $256$ & $256$ & $256$ & $256$ \\
 \hline
\end{tabular}
\caption{For each control task, below are meta-parameter configurations used for the intrinsic curiosity module (ICM), the random network distilliation (RnD), and DQN agents.}
\label{table:configs_dqn}
\begin{tabular}{l|lll|ll}
 & \multicolumn{3}{c}{\textbf{Robot-Arm Baselines}} & \multicolumn{2}{|c}{\textbf{Robot-Arm ANGC}} \\
 & \textbf{ICM} & \textbf{RnD} & \textbf{DQN} & \textbf{Controller} & \textbf{Generator} \\
 \hline 
 $\phi(\cdot)$ & ReLU  &  ReLU &  ReLU & ReLU  &  ReLU \\
 Dims & $[256, 512]$ & $[512, 256]$ & $[512,384]$ & $[512,256]$ & $[256,128]$ \\
 Rule & AdamW & Adam & Adam & Adam & Adam \\
 $\eta$ & $0.0001$ & $0.0001$ & $0.0005$ & $0.0005$ & $0.001$\\
 $\epsilon_{decay}$ & $0.97$ & $0.98$ & $0.975$ & $0.97$ & $0.97$\\
 $C$ & $100$ & $100$ & $100$ & $100$ & $100$\\
 $\gamma$ & $0.99$ & $0.99$ & $0.99$ & $0.99$ & $0.99$\\
 $N_{mem}$ & $10^6$ & $10^6$ & $10^6$ & $10^6$ & $10^6$\\
 $N_{batch}$ & $256$ & $256$ & $256$ & $256$ & $256$\\
 \hline
\end{tabular}
\caption{For the robot arm problem, below are meta-parameter configurations used for the intrinsic curiosity module (ICM), the random network distilliation (RnD), the DQN, and ANGC agents.}
\label{table:robot_configs}
\end{center}
\end{table*}

\begin{figure*}[!t] 
\centering
\begin{subfigure}[b]{.45\linewidth}
\includegraphics[width=\linewidth]{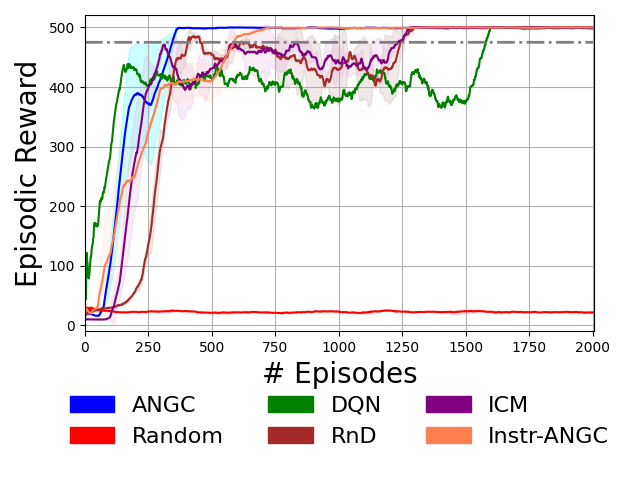}
\label{fig:cpole_mar_full}
\end{subfigure}
\begin{subfigure}[b]{.45\linewidth}
\includegraphics[width=\linewidth]{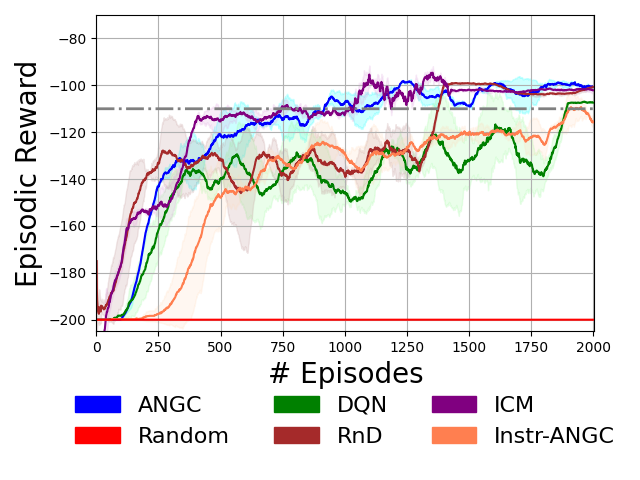}
\label{fig:mcar_mar_full}
\end{subfigure}
\begin{subfigure}[b]{.45\linewidth}
\includegraphics[width=\linewidth]{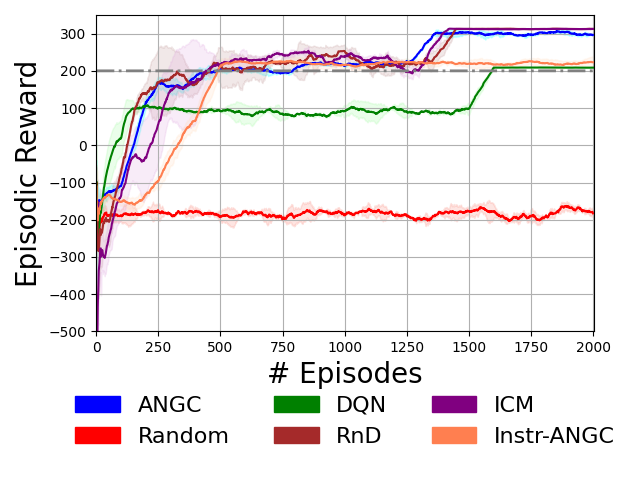}
\label{fig:lander_mar_full}
\end{subfigure}
\begin{subfigure}[b]{.45\linewidth}
\includegraphics[width=\linewidth]{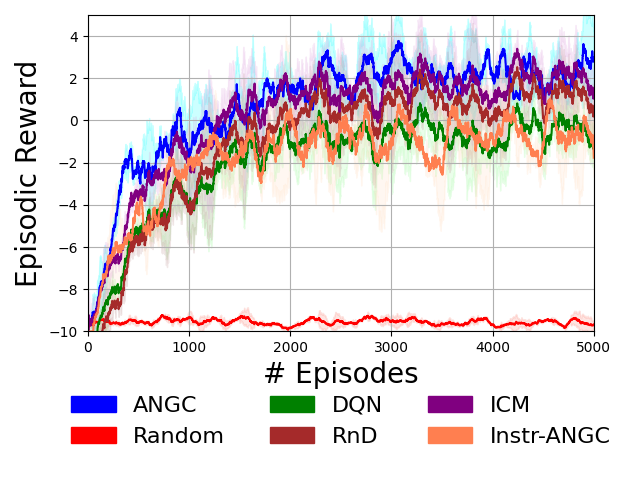}
\label{fig:robot_mar_full}
\end{subfigure}
\caption{Expanded results ($2000$ episodes) - reward curves  for ANGC and baselines (DQN, ICM, RnD, and Instr-ANGC). Mean and standard deviation over $10$ trials are plotted. Dash-dotted, horizontal (gray) lines depict the solution threshold.}
\label{fig:results_mar_full}
\end{figure*}

\subsection{Control Task Problem Descriptions}
\label{sec:control_tasks}

\paragraph{The Inverted Pendulum Problem:}
\label{sec:cpole}
In the inverted pendulum problem, also known as the cart-pole problem, the goal is to keep a pole balanced upright atop a movable cart. The state space of this problem is summarized in a 4-D vector $\mathbf{o}_t$ comprised of four values -- cart position, cart velocity, the angle $\theta$ of the pole, and the angular velocity. The agent can choose between two discrete actions -- move the cart to the left (with a fixed force value) or move the cart to the right.
The reward function for this problem yields a value of $1.0$ for every time step that the episode does not end (maximum $T = 500$). An episode ends when the angle of the pole is more than $15 ^\circ$ from vertical or when the base of the cart moves than $2.4$ units from the center of the problem space. 
This task is considered solved when an agent obtains an average cumulative reward of $475.0$ over the past $100$ episodes.

\paragraph{The Mountain Car Problem:}
\label{sec:mcar}
In this problem \cite{moore1991variable}, a car is located at the bottom of a (1-D) valley and the agent's goal is to drive the car and park at the top of the hill (to the right). The target location is marked by a flag-post at the x-coordinate value $0.5$. The state space is a 2-D vector comprised of two values -- the position of the car (along the x-axis) and the car's velocity. The agent chooses one of three discrete actions -- accelerate the car to the left, accelerate the car to the right, or do nothing.
A reward of $-1.0$ is given to the agent for every step that the episode does not end  (maximum $T = 200$) -- this encourages agents to find a way to the hilltop quickly. This task is considered solved when an agent obtains an average of $-110.0$ cumulative reward over the past $100$ episodes.

\paragraph{The Lunar Lander Problem:}
\label{sec:lander}
In the lunar lander problem, an agent is tasked with landing a rocket onto a landing pad located at the (2-D) coordinate $(0,0)$. The state space for this problem is an 8-D vector. The agent can choose between four possible discrete actions -- fire the left engine, fire the right engine, fire the rocket upwards, or do nothing. The agent receives a reward of $100$ for landing on the pad, $10$ points for each of the rocket legs that are standing, and $-0.3$ for every time step the rocket's engines are firing. 
This task is considered solved when an agent obtains an average of $200.0$ cumulative reward over the past $100$ episodes.

\paragraph{The Robot Reaching Problem:}
\label{sec:robot_arm}
In this problem, an agent manipulates a robotic arm and is tasked with reaching a ``ball'' (a red point with a location that is randomly generated every episode). The arm consist of two links where each link is $100$ pixels in length. The state space of this problem is 4-D, comprised of the target position in $x$ direction (in pixels), the target position in $y$ direction (in pixels), the current first joint position (in radians), and the current second joint position (in radians).
The agent can choose between six discrete actions -- hold the current joint angle value, increment first joint, decrement first joint, increment second joint, decrement second joint, increment joints 1 and 2, and decrement joints 1 and 2 (increments/decrements are $0.01$ radians). The agent receives a reward of $-1$ if the current distance $d$ between the tip and target position is smaller than previous distance and $+1$ if 
the value of $d$ position satisfies $-\epsilon < d < \epsilon$ ($\epsilon = 10$ pixels). An episode terminates after $100$ time steps or if the cumulative reward reaches either $-10$ or $10$.

\subsection{ANGC Agent Meta-Parameter Configurations}

\paragraph{Control Problem Configurations:}
For the classical control problems, in Table \ref{table:configs_angc}, we present several of the key values chosen (based on preliminary experimentation) for the meta-parameters of each of the two sub-models of the ANGC agent, i.e., the controller (Controller) and the generative model (Generator). Initial synaptic weight values for both were initialized by sampling a centered Gaussian with standard deviation of $0.025$.
In the table, ``Dims'' refers to the number of neurons in each latent layer of the model, e.g., $[128,128]$ means two internal layers of $128$ neurons were used, and ``Rule'' refers to the choice of the weight update rule used, such as RMSprop \cite{Tieleman2012} or Adam \cite{kingma2014adam}.
In Table \ref{table:configs_dqn}, we present the key meta-parameter settings for the primary baselines of our paper
. Note that we tuned and explored configuration settings for each baseline using a combination of heuristics and coarse grid search (the maximum number of neurons in each layer for the baselines was set to be equal to the number that would yield the exact same number of total synaptic parameters as the ANGC -- we found, empirically, that simply dumping maximum capacity to the baseline DQNs did not yield their best performance, hence why we tuned until best performance was found).

\paragraph{Robot Arm Problem Configurations:}
In Table \ref{table:robot_configs}, we present the final meta-parameter settings/configurations for the baseline DQN models as well as the ANGC for the robot arm reaching problem. Just as was done for the classical control problems, the baseline settings were carefully tuned according to a grid search (with the maximum number of neurons in each layer set in the way as in the control problems).

\subsection{Expanded Results}
In Figure \ref{fig:results_mar_full}, we present the expanded results for all four RL problems/tasks investigated in the main paper.

\section{On Modulating Synaptic Weight Updates}
\label{sec:mod_factor}

In the NGC updates provided in the main paper, we presented the update rule with an additional multiplicative modulation factor $\mathbf{M}^\ell_E$ for the generative weights and  $\mathbf{M}^\ell_E$ for the error weights. These factors, we argue, help to stabilize the model learning in the face of non-stationary streams and are formulated based on actual nonlinear synaptic dynamics. This motivation comes from neuroscience research that has demonstrated the importance of synaptic scaling (driven by competition across synapses) as a global (negative) feedback mechanism for controlling/regulating the magnitude of synaptic weight adjustments (that often rely on only locally available neural signals) \cite{turrigiano2008self,ibata2008rapid,moulin2020synaptic}.
Specifically, the rule used (to drive synaptic scaling) in our ANGC system was:
\begin{align*}
    \Delta \mathbf{W}^\ell &= \mathbf{e}^{\ell} \ast (\phi^\ell(\mathbf{z}^{\ell+1}))^T \otimes \mathbf{M}^\ell_W 
    \\
    \Delta \mathbf{E}^\ell &= \gamma_e (\Delta \mathbf{W}^\ell)^T \otimes \mathbf{M}^\ell_E
\end{align*}
where $\gamma_e$ controls the time-scale at which the error synapses are adjusted (and can generally be set to $1$ based on preliminary experimentation, though in some cases it is useful to set it to a value less than $1$ to provide additional stability by allowing the error weights to update more slowly than the generative ones, facilitating a slightly easier learning problem for NGC when the data stream is infinitely long). 
The modulation factors are themselves (locally) computed as a function of the generative weights:
\begin{align}
    \widehat{\mathbf{m}}^\ell_W &= \Sigma^{J_{\ell+1}}_{j=1} \mathbf{W}^\ell[:,j], \; \mbox{and}, \; \mathbf{m}^\ell_W = \min \bigg( \frac{\gamma_s \widehat{\mathbf{m}}^\ell_W}{\max(\widehat{\mathbf{m}}^\ell_W)}, 1 \bigg) \\
    \Delta \mathbf{M}^\ell_W &= (\mathbf{W}^\ell * 0 + 1) \otimes \mathbf{m}^\ell_W
\end{align}
where $\gamma_s = 2$ (a hyper-parameter that is also tuned) and we note that the first two formulae collapse the generative matrix to a column vector of normalized multiplicative weighting factors and the last formula converts the column vector to a tiled matrix of the same shape as $\mathbf{W}^\ell$. 
The error weight modulation factor is computed similarly to what was presented above:
\begin{align}
    \widehat{\mathbf{m}}^\ell_E &= \Sigma^{J_{\ell}}_{j=1} \mathbf{E}^\ell[:,j], \; \\
    \mathbf{m}^\ell_E &= \min \bigg( \frac{2 \widehat{\mathbf{m}}^\ell_E}{\max(\widehat{\mathbf{m}}^\ell_E)}, 1 \bigg) \\
    \Delta \mathbf{M}^\ell_E &= (\mathbf{E}^\ell * 0 + 1) \otimes \mathbf{m}^\ell_E
\end{align}
where we observe that modulation factors are computed across the pre-synaptic dimension/side of either matrix $\mathbf{W}^\ell$ or $\mathbf{E}^\ell$. Note that this multiplicative term desirably adheres to the NGC/ANGC principle of local weight adjustment (only signals immediately available are necessary to compute the modulation). 

\begin{figure*}
\centering
\begin{subfigure}[b]{.485\linewidth}
\includegraphics[width=\linewidth]{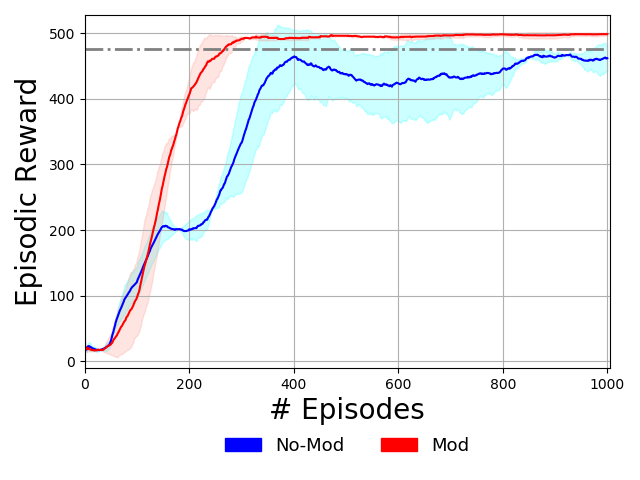}
\caption{Inverted pendulum ablation.}
\label{fig:cpole_mod}
\end{subfigure}
\begin{subfigure}[b]{.485\linewidth}
\includegraphics[width=\linewidth]{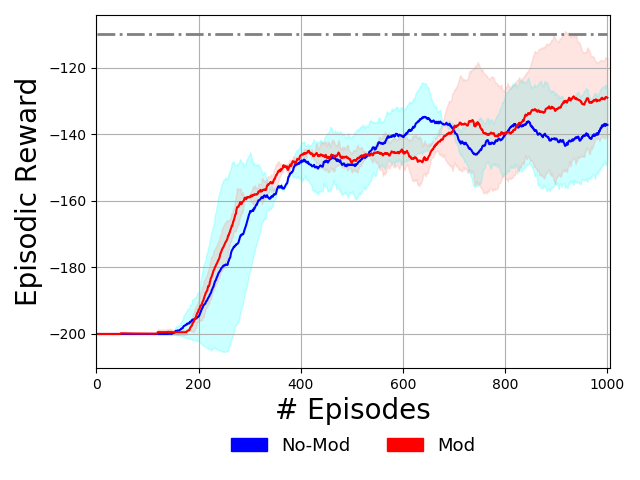}
\caption{Mountain car ablation.}
\label{fig:mcar_mod}
\end{subfigure}
\caption{Reward curves for ANGC with synaptic update modulation (Mod, red curve) and ANGC  without modulation (No-Mod, blue curve) on the inverted pendulum (left) and mountain car (right) problems. Mean and standard deviation over 10 trials are plotted. Dash-dotted, horizontal (gray) lines depict the problem solution threshold. While the gain is not always significant, in general, the modulation term appears to help the ANGC converge, in the case of the inverted pendulum, sooner to a good policy or result in higher returns earlier in the learning process (in the case of mountain car).}
\label{fig:results_modfactor}
\end{figure*}

To demonstrate the value of the multiplicative modulation term, we conducted a small ablation analysis on the three classical control benchmarks. We trained for each problem an ANGC agent for $1000$ episodes using the same configuration as the ones trained for the main experiments but under two conditions: 1) without the modulation factors (ANGC-NoMod), i.e., $\mathbf{M}^\ell_W = \mathbf{M}^\ell_E = \mathbf{1}$ (replacing each term with a matrix of ones turns off modulation), and 2) with the modulation factors $\mathbf{M}^\ell_W$ and $\mathbf{M}^\ell_E$ (ANGC-Mod). Observe in Figure \ref{fig:results_modfactor}, for both problems, a gain in average reward over time is obtained when using the modulation factor. The gain is more noticeable for the cartpole problem (see Figure \ref{fig:cpole_mod}), especially earlier in the learning phase, than for the mountain car problem (see Figure \ref{fig:mcar_mod}). However, after approximately $800$ episodes the ANGC model with the modulation factor starts to acquire consistent gains over the model without.

\section{Additional Analysis on Lunar Lander}
\label{sec:analysis}

\begin{figure*}
\centering
\begin{subfigure}[b]{.485\linewidth}
\includegraphics[width=\linewidth]{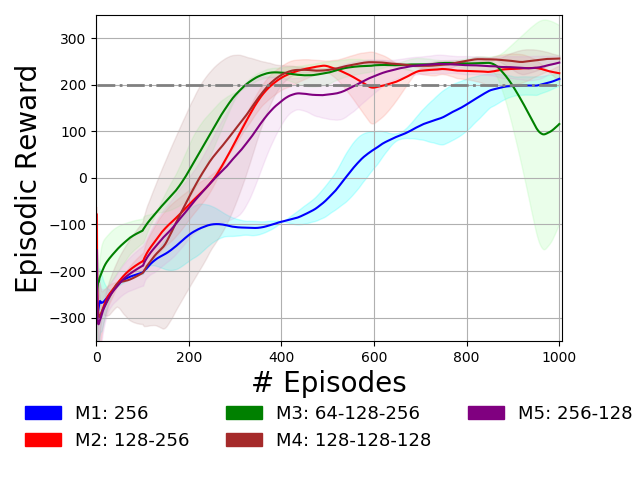}
\caption{DQN Lunar Lander Agent.}
\label{fig:lander_ddqn_analysis}
\end{subfigure}
\begin{subfigure}[b]{.485\linewidth}
\includegraphics[width=\linewidth]{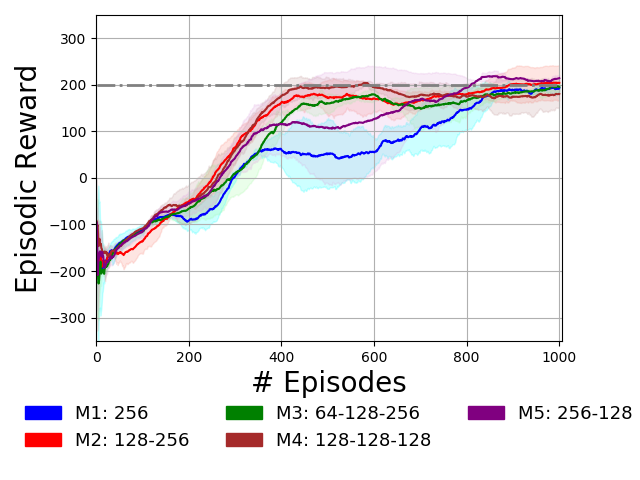}
\caption{ANGC Lunar Lander Agent.}
\label{fig:lander_angc_analysis}
\end{subfigure}
\caption{Reward curves for DQN (left) and ANGC (right) on the lunar lander problem under five different neural structures, i.e., M1 through M5. Mean and standard deviation over $10$ trials are plotted. Dash-dotted, horizontal (gray) lines depict the problem solution threshold.}
\label{fig:lander_analysis}
\end{figure*}

In Figure \ref{fig:lander_analysis}, we present a further analysis of our ANGC agent (see Figure \ref{fig:lander_angc_analysis}) against a simple DQN (see Figure \ref{fig:lander_ddqn_analysis}). In this experiment, we were interested in examining the robustness of each model's performance with respect to its underlying neural architecture/structure. We designed five DQN agents and five ANGC agents as follows:  
model \#1 (M1) had hidden dimension structure $[256]$, 
model \#2 (M2) had hidden dimension structure $[128-256]$, 
model \#3 (M3) had hidden dimension structure $[64-128-256]$,
model \#4 (M4) had hidden dimension structure $[128-128-128]$, and
model \#5 (M5) had hidden dimension structure $[256-128]$.

As observed in Figure \ref{fig:lander_analysis}, while it does appear that the categorical DQN reaches higher reward a bit earlier than the ANGC for most of the model structures (except for M1), the ANGC agents appear to yield lower variance across trials and across configurations. Notice that in the deeper, more complex model (M3, shown as the green curve in both Figure \ref{fig:lander_ddqn_analysis} and \ref{fig:lander_angc_analysis}), the average return curve takes quite a downward turn and exhibits very high variance. The ANGC under the M3 structure does not take such a downtown and furthermore exhibits very low variance. It is interesting to observe that, for the ANGC agents, all five settings appear to converge to a sound policy that reaches the lunar lander problem solution threshold, with the M5 structure (purple curve) performing best overall. This particular structure, with wider layers closer to the input consistently appeared to be an idea first choice when deciding on agent structure for all of the problems examined in this study. Nonetheless, it is more likely that the ANGC exhibits reasonably robust generalization due to the fact that it is learning not only a policy for taking discrete actions over time but also a generative world model (which drives the epistemic term of the ANGC's overall objective) that aids it in uncovering surprising parts in state space sooner in the learning process.

Another key takeaway is that inductive bias built into the ANGC (the drive to explore as a function of both epistemic and instrumental measurements of the agent's environment) appears to be important for developing adaptive agents that exhibit robust performance with respect to long term (average) reward. The AGNC generally appears to start with a higher initial return (even without training) than the DQN, which appeared to heavily dependent on model size and hidden strucutre (a result that we observed time and time again during preliminary experimentation with all DQN-based models). Only the M3 setting for the DQN yields an initial return that was similar to the ANGC agents ($\sim -200$). The DQN, while a powerful model in of itself, may or may not find a useful policy very early on its learning process and it is critical the human experimenter carefully tune its hidden structure to ensure that the agent works well across trials. For the shallow model (M1), we see that the DQN experiences a gradually increasing reward curve, finally reaching the solution threshold at about $1000$ episodes. The ANGC models, in contrast, for all five settings (M1-M5), stably reach the solution threshold all the time in just over $800$ episodes. This promising demonstration of the ANGC's (quicker) generalization ability offers some evidence why designing priors that drive more intelligent exploration are critical when dealing with challenging problem spaces such as that of the lunar lander environment.

\section{Experimental Code}
\label{sec:code}

Code will be released to reproduce the results of this paper upon acceptance of publication (the GitHub link will be provided in this appendix). Furthermore, support for applying the ANGC agent to novel/additional simulated environments will be provided to facilitate further research and extensions from the machine learning community.
Note that we have provided as part of the supplementary material, the code we did promise in the check-list for the modified robot arm reaching problem.

\textbf{Computation resources:} All experiments were performed with 128GB RAM on an intel Xeon server with 3.5GHZ processors, consisting of 4 1080Ti GPUs. Any of our models (including the ANGC/NGC ones) can easily fit into single 1080Ti GPUs -- a multi GPU setup was only used to speed up the computation. All models are coded in Tensorflow 2.2 (with eager mode) and we only use basic parallelism provided by the Tensorflow library/package to speed up the computation.
The control problem environments (including the adapted robot reaching problem) leveraged the OpenAI Gym Python package (all code was written in Python).
